\documentclass[11pt]{article}
\usepackage{lmodern}
\usepackage[T1]{fontenc}
\usepackage[margin=1in]{geometry}
\usepackage{setspace}
\usepackage{microtype}
\usepackage{amsmath, amssymb, amsthm}
\usepackage{bm}
\usepackage{caption}
\usepackage{subcaption}
\usepackage{graphicx}
\usepackage{float}
\usepackage{array}
\usepackage{multirow}
\usepackage{booktabs}
\usepackage{siunitx}
\usepackage{algorithm}
\usepackage{algpseudocode}
\usepackage{xcolor}
\usepackage{authblk}
\usepackage{tikz}
\usepackage{tabularx}
\usepackage{ltablex}
\usepackage{longtable}
\usepackage{pgfplots}
\usepackage[colorlinks=true, linkcolor=blue, citecolor=blue, urlcolor=blue]{hyperref}
\usepackage[capitalize]{cleveref}
\usepackage[numbers,sort&compress]{natbib}
\usetikzlibrary{positioning, arrows.meta, decorations.pathreplacing, calc}
\usepgfplotslibrary{groupplots}
\newcolumntype{Y}{>{\RaggedRight\arraybackslash}X}

\robustify\bfseries

\pgfplotsset{compat=1.18}
\title{\textbf{Posterior Regimes and Latent Deception: Variational Bayesian Inference in Hidden Markov Models for Sequential Fraud Detection in Financial Transactions}}

\author[1]{Joseph Uririoghene Obukofe}
\author[1]{Anthony O'Hare}
\author[2]{Chioma Sandra Dike}

\affil[1]{{\centering
    Computing Science and Mathematics, Faculty of Natural Sciences,\break
    University of Stirling, Stirling FK9 4LA, Scotland, United Kingdom\break
}}

\affil[2]{{\centering
    University of Nottingham, Nottingham, NG7 2RD, United Kingdom\break
    \normalfont\small\texttt{ouj00003@students.stir.ac.uk, anthony.ohare@stir.ac.uk, msxcd20@nottingham.ac.uk}
}}

\begin{document}
\maketitle

\begin{abstract}
    We present a three-tier progression of Hidden Markov Models: maximum-likelihood
    (Baum-Welch), variational Bayesian (VBEM), and a neural variational
    extension (Neural VBEM), that model each
    customer's transaction history as a trajectory through a small number of
    latent behavioural regimes, one of which is empirically identified as
    fraud-associated. The Neural VBEM HMM replaces the fixed Gaussian-multinomial emission
    family with a learned encoder, compressing a 741-dimensional transaction
    representation into a 64-dimensional latent space in which the VBEM HMM's
    posterior operates; a UMAP projection of this space reveals that the
    discovered regimes are not discrete clusters but ordered segments of a
    single continuous behavioural manifold, with confirmed fraud concentrated at
    its extreme. We show that the model's natural output, that is, the posterior
    probability of regime membership, is routinely mistaken for a fraud
    probability, and quantify the resulting miscalibration (the regime-membership
    interpretation error, MRIE); a corrected posterior-predictive score,
    closes most of this gap. We further distinguish
    batch (smoothed) inference, which uses look-ahead unavailable at deployment
    time, from filtered (forward-only) inference, and report both.
    On IEEE-CIS transaction data, the neural tier achieves a
    14.4$\times$ fraud enrichment in its identified regime; while its AUPRC
    trails a discriminative XGBoost baseline, we show this gap is structural
    and not incidental, and argue the model is best positioned as a
    calibrated triage and interpretability layer rather than a drop-in ranking
    replacement.
\end{abstract}

\newpage

\section{Introduction}
Card-not-present fraud detection has, for three decades, been dominated by
methods that treat each transaction as an independent observation. Rule
engines threshold hand-crafted features with no gradient of suspicion, as
logistic regression and gradient-boosted trees improve on this but still
score a transaction in isolation, and discard whatever a customer's last ten
transactions might say about the eleventh. This forms a limitation, where
card-testing attacks and synthetic-identity fraud are
all multi-transaction patterns by definition, and no
classifier that never looks at a sequence can fully characterise them.

A second and less-discussed limitation concerns uncertainty. A transaction from
a customer with three prior purchases and one from a customer with three
thousand carry very different amounts of evidence, but a standard classifier
returns a single confidence score for both, with no decomposition of how
much of that confidence is genuine signal versus how little data the model
has to go on. Under the severe class imbalance native to fraud data,
typically well under 5\% positive, this absence of calibrated uncertainty
is the setting in which naive confidence scores are most likely to mislead.

Hidden Markov Models address the first limitation directly, as they encode an
explicit hypothesis that a customer's behaviour evolves through latent
regimes, and that fraud is a regime rather than an isolated event. Classical
HMM applications to fraud detection, however, are trained by the Baum-Welch algorithm:
maximum-likelihood point estimation, and inherit the calibration
problem above, having no mechanism to express that a rarely-visited state
(which, under imbalance, includes any state associated with fraud) is
poorly estimated. This motivates treating the HMM's parameters themselves as
random variables via variational Bayesian inference, and further motivates
replacing the HMM's hand-specified Gaussian/Multinomial emission family with
a learned representation, since a fixed emission model can only ever
separate fraud from legitimate behaviour as well as the features it was
handed allow.

This paper makes four contributions. First, a
three-tier progression: Baum-Welch HMM, VBEM HMM, and Neural VBEM HMM,
in which each tier relaxes exactly one assumption of its predecessor,
letting us attribute performance changes to specific modelling choices
(\cref{sec:method}). Second, an empirical finding that the regimes the
Neural VBEM HMM discovers are not discrete
clusters but a single continuous behavioural manifold, with fraud
concentrated at one extreme rather than occupying a separate region
(\cref{sec:results}). Third, a calibration analysis showing that the
natural HMM output: the posterior probability of regime membership, is not
a fraud probability, quantifying the gap (MRIE) and providing a corrected,
provably-calibrated-in-expectation score in its place
(\cref{sec:discussion}). Fourth, an explicit separation of batch
(smoothed) and filtered (forward-only)
inference, reporting both, forming an accurate basis for any real-time claim
(\cref{sec:results}).

The remainder of the paper is organised as follows. \Cref{sec:related}
positions this work against sequential and Bayesian fraud detection
literature. \Cref{sec:method} presents the model. \Cref{sec:results}
reports model selection and predictive performance. \Cref{sec:discussion}
addresses calibration, the manifold structure, and the gap against
discriminative baselines. \Cref{sec:conclusion} concludes.

\section{Related Work}
\label{sec:related}

Logistic regression over engineered transaction features offered an
interpretable and probabilistic baseline \citep{bhattacharyyaDataMiningCredit2011},
but severe class imbalance destabilises its maximum-likelihood estimates,
and its linearity limits the interaction structure it can capture.
Gradient-boosted trees, XGBoost \citep{chenXGBoostScalableTree2016} and
LightGBM \citep{keLightGBMHighlyEfficient2017} in particular, have since
come to dominate tabular fraud benchmarks by capturing high-order feature
interactions without manual specification; XGBoost serves as our primary
discriminative baseline in \cref{sec:results} for this reason. What
every method in this family shares, independent of its individual
strengths, is a transaction-level unit of analysis, as each row is scored on
its own, with no mechanism for one transaction to inform the model's
belief about another.

Jurgovsky et al.\ \citep{jurgovskySequenceClassificationCreditcard2018}
provided an early systematic evaluation of LSTMs
\citep{hochreiterLongShortTermMemory1997} for credit card fraud, and
transformer-based architectures have since been applied to the same
setting \citep{huangTabTransformerTabularData2020}. Both families improve
on transaction-independent scoring, but both remain discriminative, where a
customer's behavioural state is encoded only implicitly, as a hidden
vector optimised for a downstream fraud label, thus with no explicit
representation of what regime the customer is currently in, no transition
structure between regimes, and no calibrated uncertainty over the state
itself. A Hidden Markov Model, by contrast, treats the latent regime as an
object with its own distribution and its own learned dynamics.

Hidden Markov Models bring this explicit latent-state structure to
sequential fraud detection, and Srivastava et al.\
\citep{srivastavaCreditCardFraud2008} provide
the direct precedent for applying one here, using discretised transaction
amount and merchant category as emission symbols. Classical HMM fraud
detection, however, is fit by Baum-Welch, an EM procedure that returns
point estimates with no posterior over parameters, forming a liability for
a state that, under typical fraud prevalence, is visited rarely enough
that its maximum-likelihood estimate is built on very little evidence.
Exact Bayesian treatment of an HMM's parameters is available via MCMC
\citep{scottBayesianMethodsHidden2002}, but blocked Gibbs sampling over the
full state sequence does not scale to a dataset of this size or to
real-time scoring. Variational Bayesian EM
\citep{bealVariationalAlgorithmsApproximate2003} resolves this, where
closed-form, conjugate posterior updates over the transition matrix and
emission parameters, retaining calibrated uncertainty without MCMC's
computational cost, which is the inference framework the VBEM HMM tier of
this paper adopts directly.

The VBEM HMM, like its maximum-likelihood predecessor, still inherits a
fixed parametric emission family applied to the raw feature space, and
cannot recover interactions or non-linear structure that family was not
designed to express. The closest prior work to our solution is the Deep
Markov Model \citep{krishnanStructuredInferenceNetworks2016}, which also
pairs a Markovian latent-state model with a neural network to escape a
fixed emission family. The two approaches invert each other's direction, where a
Deep Markov Model defines the likelihood of observed data given
the hidden representation via a neural decoder mapping
latent states outward to observations, which requires a generative
reconstruction loss and does not compose directly with exact
forward-backward inference. The Neural VBEM HMM instead defines a neural
encoder mapping observations inward to a latent embedding, and places the
VBEM's Gaussian emission in that learned space, preserving the full
forward-backward recursion, the Dirichlet and Normal-Inverse-Wishart
posteriors, and closed-form inference exactly as in the VBEM tier, while
letting the encoder learn what the emission geometry should be as opposed to
fixing it by hand.

\section{The Neural Variational Bayesian Hidden Markov Model}
\label{sec:method}

\subsection{Problem Setup}
\label{sec:problem-setup}

A labelled transaction dataset is reorganised into one chronologically
ordered sequence per customer, $\mathbf{X}_u = (\mathbf{x}_{u,1}, \ldots,
    \mathbf{x}_{u,T_u})$, where each $\mathbf{x}_{u,t} \in \mathbb{R}^F$
partitions into $D$ continuous and $J$ categorical features ($F = D + J =
    741$ in our setting). Fraud labels are used only to evaluate the model and
to align its discovered latent structure with observed outcomes after
fitting; training itself is unsupervised. Sequences shorter than $T_{\min}
    = 5$ carry too little evidence for a customer-specific posterior to be
meaningful and are excluded from direct sequence-level fitting, handled
instead by the hierarchical prior extension in \cref{sec:hierarchical}.

The central modelling claim is that a customer's observable behaviour at
any point is generated by one of a small number of unobserved behavioural
regimes, that these regimes persist over consecutive transactions rather
than switching arbitrarily, and that a transaction is fraudulent as a
function of which regime produced it. A Hidden Markov Model is the natural approximation to
assess for this claim, with hidden regimes evolving under a first-order
Markov process, each emitting a transaction through a regime-specific distribution.

\subsection{The Hidden Markov Model Framework}
\label{sec:hmm-framework}

At every step $t$, customer $u$'s account occupies one of $K$ latent
states $q_{u,t} \in \{1, \ldots, K\}$, governed by an initial distribution
$\bm{\pi}$ and a transition matrix $A$, with each transaction emitted from
a state-specific distribution $B_k(\cdot)$. The joint probability of an
observed sequence and its hidden path is given as

\[
    p(\mathbf{X}_u, \mathbf{q}_u \mid \bm{\theta}) = \pi_{q_{u,1}}
    B_{q_{u,1}}(\mathbf{x}_{u,1}) \prod_{t=2}^{T_u} A_{q_{u,t-1},q_{u,t}}
    B_{q_{u,t}}(\mathbf{x}_{u,t}),
\]

and the marginal likelihood, obtained by summing over all $K^{T_u}$
hidden paths, is intractable to evaluate directly. The forward-backward
algorithm resolves this via dynamic programming
\citep{rabinerTutorialHiddenMarkov1989}; in log space,

\begin{align}
    \log \alpha_t(k) & = \log B_k(\mathbf{x}_{u,t}) + \log \sum_{j=1}^K
    \exp\bigl(\log \alpha_{t-1}(j) + \log A_{jk}\bigr),                 \\
    \log \beta_t(k)  & = \log \sum_{j=1}^K \exp\bigl(\log A_{kj} + \log
    B_j(\mathbf{x}_{u,t+1}) + \log \beta_{t+1}(j)\bigr),
\end{align}

initialised at $\log \alpha_1(k) = \log \pi_k + \log B_k(\mathbf{x}_{u,1})$
and $\log \beta_{T_u}(k) = 0$.

The emission $B_k$ factorises across the continuous and categorical
feature blocks under a conditional-independence assumption
\citep{rabinerTutorialHiddenMarkov1989,bishopPatternRecognitionMachine2007},
with a diagonal-covariance Gaussian over the continuous block and an
independent categorical distribution per discrete feature:

\[
    \log B_k(\mathbf{x}_{u,t}) = \log \mathcal{N}(\mathbf{x}_{u,t}^{\mathrm{cont}}
    \mid \bm{\mu}_k, \bm{\Sigma}_k) + \sum_{j=1}^J \log
    \pi^{(j)}_{k, x^{\mathrm{cat}}_{u,t,j}}.
\]

The diagonal restriction reduces the covariance parameter count from
quadratic to linear in $D$ ($83{,}028$ versus $407$ parameters per state
at $D = 407$), which matters for a state that, under $3.5\%$
fraud prevalence, has comparatively little data to estimate from.

\subsection{From Point Estimates to Variational Bayesian Inference}
\label{sec:vb-inference}

The Baum-Welch baseline fits $\bm{\theta}$ by Expectation-Maximisation,
alternating the forward-backward recursion above with closed-form
maximum-likelihood updates
\citep{baumStatisticalInferenceProbabilistic1966,baumMaximizationTechniqueOccurring1970,dempsterMaximumLikelihoodIncomplete1977};
we do not re-derive these standard updates here (full forms in
\cref{app:derivations}). The resulting point estimates carry no notion of
their own uncertainty, forming a liability for a rarely-visited state
whose maximum-likelihood parameters are fit to very little effective
evidence with no mechanism to signal that fact.

The VBEM tier addresses this by placing conjugate priors over every
parameter and learning posteriors, specifically, Dirichlet
over $\bm{\pi}$ and each row of $A$, Normal-Inverse-Wishart (NIW) over
$(\bm{\mu}_k, \bm{\Sigma}_k)$
\citep{bealVariationalAlgorithmsApproximate2003,murphyConjugateBayesianAnalysis2007}.
The transition prior is deliberately ``sticky'', biased toward
self-transition to encode that behavioural regimes persist across
consecutive transactions \citep{foxStickyHDPHMMApplication2011}:

\[
    \alpha_{ii} = \alpha_0 + \kappa, \qquad \alpha_{ij} = \alpha_0 \ (i \neq
    j), \qquad \kappa > 0.
\]

Under the mean-field assumption $q(\bm{\theta}) = \prod_i
    q_i(\theta_i)$, the intractable true posterior is replaced by the
variational family that maximises the evidence lower bound,

\[
    \mathcal{L}(q) = \mathbb{E}_q\!\left[\log \frac{p(\mathbf{X},
            \bm{\theta})}{q(\bm{\theta})}\right],
\]

which, unlike the Bayesian Information Criterion used for Baum-Welch model
selection \citep{schwarzEstimatingDimensionModel1978}, penalises
complexity in proportion to how far each parameter's posterior has
actually moved from its prior and not by a fixed count of free
parameters; the full expansion of $\mathcal{L}(q)$ into its Dirichlet and
NIW Kullback-Leibler terms is given in \cref{app:derivations}. We train
candidate models for $K \in \{2, \ldots, 10\}$ and select $K^*$ by the
converged ELBO. The reported representative orders also depend on label-based
regime diagnostics, described in \cref{sec:predictive-performance};
they are not presented as simply the global BIC minima or ELBO maxima.

\subsection{The Neural Extension}
\label{sec:neural-extension}

Both preceding tiers still apply a fixed Gaussian/categorical emission
family directly to the raw $F$-dimensional feature space, which cannot
represent multimodal fraud subtypes occupying disjoint regions of that
space, nor interactions between features that only jointly signal fraud.
We address this by learning the emission geometry instead of fixing it. A
neural encoder $f_\theta : \mathbb{R}^F \to \mathbb{R}^{d_z}$ maps each
transaction to a $d_z$-dimensional latent embedding, and the VBEM Gaussian
emission is placed on $\mathbf{z}_{u,t} = f_\theta(\mathbf{x}_{u,t})$
instead:

\[
    p(\mathbf{z}_{u,t} \mid q_{u,t} = k) = \mathcal{N}(\mathbf{z}_{u,t} \mid
    \bm{\mu}_k^z, \bm{\Sigma}_k^z).
\]

The encoder processes continuous and categorical features through separate pathways,
using an MLP over layer-normalised continuous features and per-feature embedding
tables for the categorical block \citep{baLayerNormalization2016,gorishniyRevisitingDeepLearning2023}.
The resulting continuous and categorical representations are then concatenated,
passed through a shared linear transformation, normalised, transformed using
a GELU activation, and finally projected into the latent representation space of dimension $d_z$.
Architecture and hyperparameter detail (hidden dimensions, the
embedding-size heuristic, dropout) is given in
\cref{app:implementation}; we treat $d_z$ as a hyperparameter selected
empirically over $\{32, 64, 128\}$.

\textbf{Decoupled training.} Our first attempt trained the encoder and the
VBEM posteriors jointly, alternating an encoder gradient step with each EM
iteration and using the current posteriors as the encoder's training
target. Under $96.5\%$--$3.5\%$ class imbalance this produced a specific and
reproducible failure, where the dominant legitimate state's larger effective
count dominated every encoder gradient, pulling all embeddings, fraud
included, toward the legitimate centroid. The VBEM mechanic responded by
concentrating posteriors further on that state, amplifying the imbalance
for the next encoder update. This closed feedback loop, which we term
``occupancy collapse'', proved irreversible once underway and was not
fixed by loss reweighting or reseeding. We resolve this by fully
decoupling the two components. The encoder is first trained to
convergence alone, as a class-weighted binary classifier with a temporary
linear head,

\[
    \mathcal{L}(\theta, w, b) = -\frac{1}{N}\sum_{u,t}\bigl[\omega\,
        y_{u,t}\log\hat y_{u,t} + (1-y_{u,t})\log(1-\hat y_{u,t})\bigr], \qquad
    \hat y_{u,t} = \sigma(w^\top \mathbf{z}_{u,t} + b),
\]

with $\omega = N_-/N_+ \approx 27.6$ equalising fraud and legitimate
gradient contribution. The head is then discarded, the encoder frozen, and
every transaction re-encoded once. The VBEM runs its standard and
unmodified coordinate ascent on these fixed embeddings, receiving no
further gradient from the encoder and contaminating no further encoder
update. Because model-order selection compares the ELBO across $K$, and
each $K$ here induces its own independently pretrained encoder and
therefore a different observation space, ELBO values are not comparable
across $K$ for this tier; we instead select $K^*$ by held-out
log-likelihood on a customer-level validation split, treating the
training-set ELBO as a secondary convergence diagnostic only.

This architecture is closely related to the Deep Markov Model
\citep{krishnanStructuredInferenceNetworks2016} (\cref{sec:related}), but
inverts its direction, as rather than a neural decoder mapping latent states
outward to a reconstruction, we learn an encoder mapping observations
inward, and place the emission there, allowing the
forward-backward recursion, the Dirichlet and NIW posteriors, and
closed-form VBEM inference carry over from the previous tier unchanged.

\subsection{Hierarchical Prior and Cold-Start Scoring Extension}
\label{sec:hierarchical}

A pooled HMM can score a new customer from the first observed transaction: initialise its latent state
distribution with the shared $\bm{\pi}$ then apply the filtered updates in
\cref{sec:inference-scoring}. The customer has an individual belief state,
while the emission and transition parameters remain shared. This
requires neither a separate per-customer parameter posterior nor an arbitrary minimum history
length at inference.

As transactions accumulate, the posterior shifts continuously from this
population prior toward a customer-specific one, giving a smooth
cold-start-to-warm-start transition with no explicit regime switch. The
same construction applies unchanged to the Neural VBEM tier, over the
encoder-induced latent statistics rather than the raw feature statistics.

A hierarchical model with customer-specific emission parameters would be a separate extension. It
would require explicit population hyperpriors, customer-level likelihoods, aligned state definitions,
and an inference scheme. Averaging posterior location and scale matrices alone does not establish a
moment-matched NIW distribution. No separate cold-start evaluation is reported here, and the
pooled-model results are not offered as validation of such a hierarchical extension.

\subsection{Inference and Fraud Scoring}
\label{sec:inference-scoring}

A fitted model is a single specification, estimated once across the
pooled training population, as it does not differ across customers, only the
latent path $q_u$ each customer's history induces through it does. The
fraud-associated state is identified once, after convergence, as $k^* =
    \arg\max_k \eta_k$, where $\eta_k$ is the empirical fraud rate among
transactions assigned in expectation to state $k$.

We distinguish three inference regimes, which return the same underlying
object under different information constraints. \emph{Batch} (smoothed)
inference uses the complete sequence, conditioning each transaction's
score on both past and future observations via $\gamma_t(k) =
    p(q_{u,t}=k \mid \mathbf{X}_u, \bm\theta)$, giving
$\mathcal{F}_{\mathrm{batch}}(\mathbf{x}_{u,t}) = \gamma_t(k^*)$.
\emph{Retrospective filtered} inference replays a stored historical
sequence but withholds every transaction after $t$ when scoring $t$, using
the forward pass alone, forming an offline approximation of what a
live system would have known, and the gap between it and the batch score
at the same transaction is itself a diagnostic for how much of a
transaction's apparent fraud evidence comes from what happens afterward.
\emph{Real-time} inference is the true live case, as transaction $t+1$
arrives after $t$ has already been scored and cannot influence it. We
formalise the object real-time scoring depends on as the transaction
consequence, $\mathbf{c}_{u,t} \in \Delta^{K-1}$, which coincides
with the filtered posterior, $\mathbf{c}_{u,t}(k) \equiv
    \gamma^{\mathrm{filt}}_t(k)$, updated one transaction at a time rather
than recomputed from a stored sequence:

\[
    \mathbf{c}_{u,t} = \frac{(A^\top \mathbf{c}_{u,t-1}) \odot \mathbf{b}_t}
    {\mathbf{1}^\top\bigl[(A^\top \mathbf{c}_{u,t-1}) \odot
            \mathbf{b}_t\bigr]}, \qquad \mathbf{c}_{u,0} := \bm{\pi},
\]

with $\mathbf{b}_t$ the emission likelihood vector at $t$. Because
$\mathbf{c}_{u,t}$ must persist between a customer's transactions,
arbitrarily far apart in real time and potentially across separate server
instances, it is stored as a small per-customer state external to the
model parameters $\bm\theta$, which remain a single static, shared
artefact.

\textbf{Regime membership is not a fraud probability.} $\gamma_t(k^*)$, in
any of the three inference regimes above, is the posterior probability that a
transaction belongs to state $k^*$. It is not the probability that the
transaction is fraudulent, and treating the two as interchangeable
silently assumes $\eta_{k^*} = 1$, which never holds for a state built
from real and mixed data. The best fraud-probability estimate available from
the model's posterior alone is instead the full mixture, weighted by each
state's own empirical fraud rate:

\[
    \mathcal{F}_{\mathrm{corrected}}(\mathbf{x}_{u,t}) = \sum_{k=1}^K
    \gamma_t(k) \cdot \eta_k.
\]

This assumes within-state fraud homogeneity: that every transaction
assigned to state $k$ shares that state's average fraud rate $\eta_k$.
Given that the posterior over states is the entirety of what the model
knows about a transaction, this is the best estimate the
available information supports. It has a further
property worth stating, that its population mean is,
by construction, the true population fraud rate,

\[
    \frac{1}{N}\sum_{u,t} \mathcal{F}_{\mathrm{corrected}}(\mathbf{x}_{u,t}) =
    \sum_{k=1}^K \eta_k \cdot \frac{N_k}{N} = \sum_{k=1}^K \frac{F_k}{N} =
    \frac{1}{N}\sum_{u,t} y_{u,t} = \bar y,
\]

where $N_k = \sum_{u,t}\gamma_{u,t}(k)$ and $F_k =
    \sum_{u,t}\gamma_{u,t}(k)\,y_{u,t}$ are the effective state occupancy and
fraud counts already used to define $\eta_k = F_k/N_k$, and $\sum_k F_k =
    \sum_{u,t} y_{u,t}$ follows from $\sum_k \gamma_{u,t}(k) = 1$. This
identity holds regardless of how good the underlying regime structure is,
and $\gamma_t(k^*)$ has no analogue of it. Both scores are defined
identically under batch or filtered posteriors; \cref{sec:results}
reports both.

\section{Results}
\label{sec:results}

\subsection{Data and Evaluation Scope}
\label{sec:data-scope}

The data source is the IEEE-CIS Fraud Detection competition
\citep{ieeeCIS2019}. The supplied labelled transaction file contains
590,540 rows, 20,663 fraud labels, and 13,553 distinct \texttt{card1}
values. Transactions are left-joined to identity records on
\texttt{TransactionID}. The preprocessing code randomly splits unique
\texttt{card1} values 85/15 with seed 42. Saved training and evaluation
partitions contain 504,824 and 85,716 rows, respectively, and have no
\texttt{card1} overlap. Their time ranges overlap: this is not a
chronological holdout. The grouping field is a proxy rather than an
established customer identifier.

HMM sequence assembly excludes groups shorter than five transactions,
leaving 493,401 training rows in 5,527 groups and 83,791 evaluation rows
in 985 groups. The saved evaluation labels contain 2,792 frauds. The
baseline result tables instead contain 85,716 evaluation rows and 2,856
frauds. The neural training script holds out another 15\% of the retained
training groups for model-order diagnostics; encoder pretraining makes
an additional stratified 15\% transaction-level split within its fitting
pool for early stopping. Preprocessing is fitted before these inner
splits. Thus the internal validation preprocessing is not independently
fitted, and the encoder stopping split can share groups.

\subsection{Model Selection}
\label{sec:model-selection}

For each tier and $K \in \{2, \ldots, 10\}$, a candidate model order is
eligible only if (1) the emission-based and posterior-based
fraud-state assignments agree, (2) the agreed state carries at least
$0.5\%$ of the effective observations, and (3) its fraud rate exceeds the
population base rate ($\approx 3.5\%$). Among eligible candidates, $K^*$ is
selected by fraud-state enrichment, with a parsimony tolerance of $\delta
    = 0.5$ preferring a simpler model unless a more complex one exceeds it by
more than $50\%$; ties fall to the higher empirical fraud rate.

Criterion (1) does not apply to the Neural VBEM tier, whose latent coordinates carry
no fixed semantic meaning for an emission-based proxy to be computed
against, as every candidate order for that tier proceeds directly to ranking.
Engineered features outperformed processed features at every tier and
model order tested; only engineered results are reported below, with the
processed-feature comparison in \cref{app:additional-diagnostics}. The
full per-$K$, per-tier trail behind the selections in
\cref{tab:model-selection-summary} is given in
\cref{app:extended-model-selection}.

\begin{table}[H]
    \centering
    \small
    \setstretch{1.3}
    \setlength{\aboverulesep}{0.7ex}
    \setlength{\belowrulesep}{0.7ex}
    \setlength{\tabcolsep}{5pt}
    \caption{Selected model order and fraud-state diagnostics for each HMM
        tier under batch posteriors for engineered features.}
    \label{tab:model-selection-summary}
    \begin{tabular}{lccccc}
        \toprule
        \textbf{Model}             & \boldmath{$K^*$} & \textbf{Occupancy} & \textbf{Fraud rate} & \textbf{Enrichment} \\
        \midrule
        Baum-Welch HMM             & 8                & 1.77\%             & 19.4\%              & $5.48\times$        \\
        VBEM HMM                   & 7                & 3.78\%             & 25.3\%              & $7.16\times$        \\
        Neural VBEM HMM ($d_z=64$) & 9                & 3.8\%              & 50.9\%              & $14.43\times$       \\
        \bottomrule
    \end{tabular}
\end{table}

\subsection{Predictive Performance}
\label{sec:predictive-performance}

Performance is assessed on three complementary metrics: the Area Under
the Precision-Recall Curve (AUPRC) for ranking under severe class
imbalance, the Kolmogorov-Smirnov (KS) statistic for the maximum
separation between fraud and legitimate score distributions
\citep{masseyKolmogorovSmirnovTestGoodness1951}, and the Expected
Calibration Error (ECE) for agreement between predicted and observed
fraud frequencies \citep{guoCalibrationModernNeural2017,naeiniObtainingWellCalibrated2015}.

\Cref{tab:hmm-progression} reports all three metrics under three
successive scoring conditions for each tier: the raw regime-membership
score $\gamma_t(k^*)$ under batch (smoothed) posteriors, as earlier
reported; the corrected score $\mathcal{F}_{\mathrm{corrected}}$
(\cref{sec:inference-scoring}) under the same batch posteriors, isolating
the effect of the calibration fix alone; and the corrected score under
filtered (forward-only) posteriors.

\begin{table}[H]
    \centering
    \small
    \setstretch{1.3}
    \setlength{\aboverulesep}{0.7ex}
    \setlength{\belowrulesep}{0.7ex}
    \setlength{\tabcolsep}{5pt}
    \caption{HMM tier predictive performance under three scoring conditions.
        Raw/batch is $\gamma_t(k^*)$, smoothed posteriors;
        corrected/batch and corrected/filtered use
        $\mathcal{F}_{\mathrm{corrected}} = \sum_k \gamma_t(k)\eta_k$
        (\cref{sec:inference-scoring}) under smoothed and forward-only posteriors
        respectively.}
    \label{tab:hmm-progression}
    \begin{tabular}{llccc}
        \toprule
        \textbf{Model} & \textbf{Score}       & \textbf{AUPRC} & \textbf{KS} & \textbf{ECE} \\
        \midrule
        \multirow{3}{*}{Baum-Welch HMM}
                       & raw / batch          & 0.0775         & 0.2667      & 0.0414       \\
                       & corrected / batch    & 0.0853         & 0.3883      & $\approx 0$  \\
                       & corrected / filtered & 0.0853         & 0.3883      & $\approx 0$  \\
        \midrule
        \multirow{3}{*}{VBEM HMM}
                       & raw / batch          & 0.0667         & 0.2151      & 0.0685       \\
                       & corrected / batch    & 0.0806         & 0.2841      & $\approx 0$  \\
                       & corrected / filtered & 0.0806         & 0.2841      & $\approx 0$  \\
        \midrule
        \multirow{3}{*}{Neural VBEM HMM}
                       & raw / batch          & 0.1987         & 0.4714      & 0.0487       \\
                       & corrected / batch    & 0.1918         & 0.4684      & 0.000115     \\
                       & corrected / filtered & 0.1932         & 0.4686      & 0.000114     \\
        \bottomrule
    \end{tabular}
\end{table}

The ECE collapse under correction is the calibration identity from
\cref{sec:inference-scoring} showing up empirically, because
$\mathcal{F}_{\mathrm{corrected}}$'s population mean equals the true
fraud rate by construction, and its values cluster narrowly around that
rate across states, almost the entire evaluation set falls into one or
two calibration bins whose average matches the population rate almost
exactly. AUPRC and KS move by a smaller amount under
correction, since $\mathcal{F}_{\mathrm{corrected}}$ is not a monotonic
transformation of $\gamma_t(k^*)$ and can re-rank transactions whose belief
mass is spread across several moderately fraud-associated states, rather
than concentrated in $k^*$ alone. The raw/batch row is the only one of
the three with bootstrap validation to date, as 95\% confidence intervals
over 1{,}000 resamples are non-overlapping across all three tiers on both
AUPRC and KS (Neural VBEM AUPRC $[0.187, 0.211]$ against Baum-Welch
$[0.072, 0.085]$ and VBEM $[0.062, 0.072]$), and a paired Wilcoxon
signed-rank test finds the Neural VBEM ahead of both raw-feature tiers on every one of the
    1{,}000 resamples ($W=0$, $p<10^{-100}$) on both AUPRC and KS.

One counter-intuitive result from that same test is,
on the raw/batch ECE specifically, Baum-Welch is
significantly better calibrated than Neural VBEM ($\Delta=+0.007$,
$p<10^{-100}$), highlighting a consequence of its much smaller fraud-state occupancy
(1.77\% against 3.8\%) exposing less of the population to the
regime-membership gap and not of any superior probability estimation.

\Cref{tab:baseline-comparison} places the Neural VBEM HMM against two
non-sequential baselines: Isolation Forest, an unsupervised anomaly
detector \citep{liuIsolationForest2008}, and XGBoost
\citep{chenXGBoostScalableTree2016}, the strongest tuned discriminative
classifier available on this task (\cref{sec:related}).

\begin{table}[H]
    \centering
    \small
    \setstretch{1.3}
    \setlength{\aboverulesep}{0.7ex}
    \setlength{\belowrulesep}{0.7ex}
    \setlength{\tabcolsep}{5pt}
    \caption{Predictive performance against non-sequential baselines on
        held-out evaluation set. Neural VBEM figure is the raw/batch score,
        $\gamma_t(k^*)$; see note below.}
    \label{tab:baseline-comparison}
    \begin{tabular}{lccc}
        \toprule
        \textbf{Model}   & \textbf{AUPRC} & \textbf{KS}    & \textbf{ECE}   \\
        \midrule
        Isolation Forest & 0.098          & 0.365          & 0.458          \\
        Neural VBEM HMM  & 0.199          & 0.471          & 0.049          \\
        XGBoost          & \textbf{0.514} & \textbf{0.563} & \textbf{0.019} \\
        \bottomrule
    \end{tabular}
\end{table}

The table is not a matched comparison: it combines different evaluation
populations, feature sets, and selection procedures. XGBoost and Isolation
Forest receive all columns except the label in their engineered tables,
whereas the HMMs explicitly route continuous and categorical features and
exclude identifying/time metadata. Both baseline searches use five-fold
shuffled stratified transaction-level cross-validation, 100 Bayesian-search
iterations, and average precision for selection. Isolation Forest fitting
is unsupervised, but its hyperparameter selection uses fraud labels.
Its displayed probability-like score is a sigmoid of the negated decision
function, not a learned probability calibration. Its ECE must therefore
not be read as directly comparable calibration performance. A matched
comparison requires common transaction identities, causal features,
separate selection and calibration data, and repaired preprocessing.

\subsection{Exploratory Latent-Space Visualisation}
\label{sec:latent-manifold}

\begin{figure}[H]
    \centering
    \includegraphics[width=\linewidth]{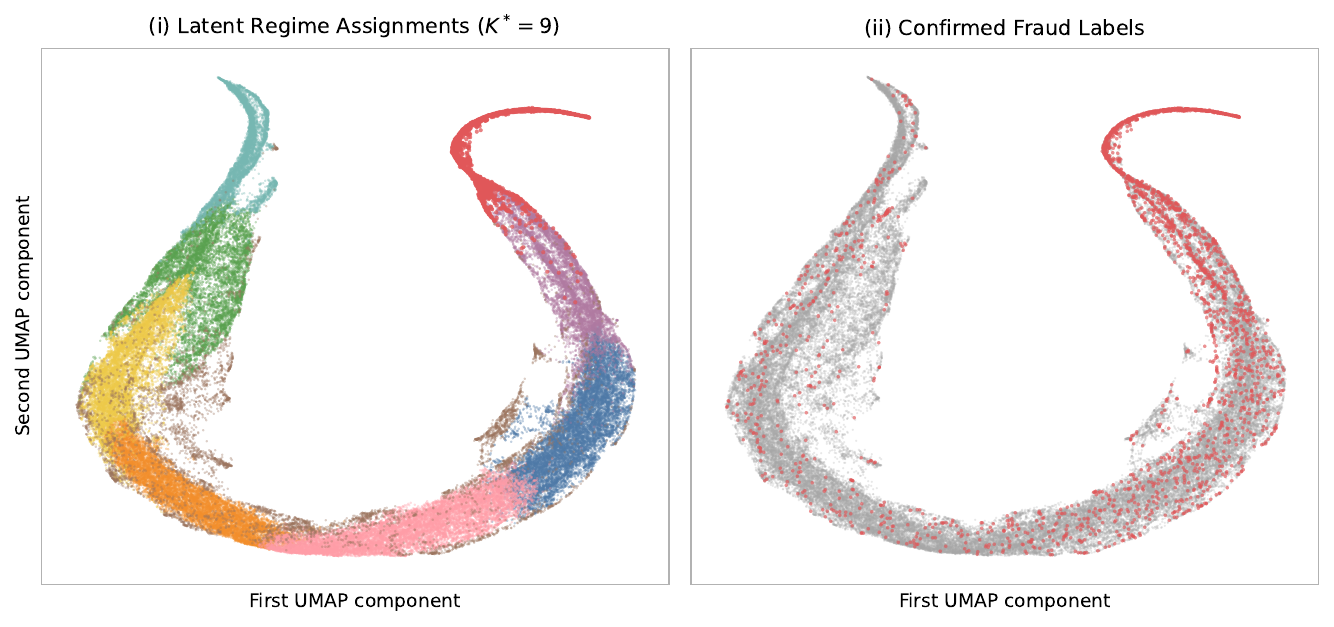}
    \caption{UMAP projection of the Neural VBEM HMM ($d_z=64$, $K^*=9$)
        latent embeddings. Panel (i): each transaction coloured by its hard
        state assignment, fraud-associated state $k^*=4$ highlighted in red.
        Panel (ii): the same projection coloured by ground-truth fraud label.
        Confirmed fraud concentrates at the same terminal tip, density
        diminishing continuously toward the legitimate-dominated body.}
    \label{fig:nvbem_umap}
\end{figure}

The Neural VBEM's decoupled training exposes an inspectable latent
representation, $\mathbf{z}_{u,t} = f_\theta(\mathbf{x}_{u,t}) \in
    \mathbb{R}^{64}$, compressed from the 741 raw features by an encoder
trained to separate confirmed fraud from legitimate transactions before
the VBEM ever runs. \Cref{fig:nvbem_umap} shows a UMAP projection
\citep{mcinnesUMAPUniformManifold2020} of this space.

The nine inferred regimes are ordered,
contiguous segments of a single continuous manifold,
transitioning from legitimate-spending
regimes at the body toward the fraud-associated state $k^*$ at its
extreme. Confirmed fraud concentrates at that tip, with density
falling off continuously along the manifold as opposed to switching at a
boundary. The $42.1\%$ of confirmed fraud not
captured by $k^*$ appear scattered through the
manifold's body in panel (ii), interspersed with legitimate transactions
at geometrically similar latent positions, behaviourally
indistinguishable from them in the representation the encoder has
learned. This forms a limit of the model, and is the geometric explanation for why the
AUPRC gap against XGBoost in \cref{tab:baseline-comparison} is structural, as
a classifier's transaction-level ranking is not bounded by this
representation's geometry the way regime-membership scoring is.

\section{Discussion}
\label{sec:discussion}

\subsection{Evaluation Asymmetry: HMMs vs Discriminative Classifiers}
\label{sec:evaluation-asymmetry}

XGBoost's training objective explicitly rewards ranking every confirmed
fraud above every confirmed legitimate transaction across the full
feature space, which is what AUPRC measures; classifier and
metric are aligned by design. The HMM family's objective, the
ELBO, is dominated by the $96.5\%$ legitimate transaction mass and
carries no such instruction. Concretely, the fraud-associated regime
under the selected Neural VBEM configuration holds roughly $3{,}176$
transactions at $50.9\%$ purity, so about $1{,}617$ of the dataset's
$2{,}792$ confirmed frauds sit inside it ($57.9\%$); the remaining
$1{,}175$ ($42.1\%$) receive low scores because their posterior mass
lies in regimes that resemble legitimate behaviour, dragging the
precision-recall curve down at every threshold where they rank below
legitimate transactions, independent of how well the model performs on
the fraud it does capture confidently.

\subsection{Regime Membership, Fraud Probability, and the Calibration Gap}
\label{sec:calibration-gap}

As an illustration using the archived training rate, if
$\gamma_t(k^*)=0.90$ and $\widehat\eta_{k^*}=0.509$, the selected
state contributes $0.4581$ to the mixture score. The complete score is

\[
    S_t=0.4581+\sum_{k\ne k^*}\gamma_t(k)\widehat\eta_k.
\]

Ignoring the other states gives a discrepancy of $0.4419$, or
$44.19$ percentage points, from the membership score. It is not
generally the discrepancy from the complete fraud-probability estimate.
For the Mean Regime Interpretation Error (MRIE), define

\[
    \mathrm{MRIE}=\frac1N\sum_{u,t}\gamma_{u,t}(k^*)
    (1-\widehat\eta_{k^*})
    =\mathrm{o}_{k^*}(1-\widehat\eta_{k^*}).
\]

This is a diagnostic of the gap between raw membership and the selected
state's contribution alone. The average signed gap from the full mixture is

\[
    \frac1N\sum_{u,t}\bigl[\gamma_{u,t}(k^*)-S_{u,t}\bigr]
    =\mathrm{MRIE}-\sum_{k\ne k^*}\mathrm{o}_k\widehat\eta_k.
\]

Neither quantity is an absolute calibration error. Even a constant
prevalence predictor can match the average outcome without useful
discrimination. The mixture is bounded between the smallest and largest
state rates, but this bound and the in-sample mean identity do not imply
calibration on new customers or under distribution shift.

\subsection{Limitations}
\label{sec:limitations}

The gap between corrected/batch and corrected/filtered scoring was small
for every tier tested (\cref{tab:hmm-progression}), meaning a live
deployment can expect scores close to what this paper's offline
evaluation reports, rather than to the batch numbers this literature has
earlier reported alone. Separately, the eligibility framework's
thresholds, the $0.5\%$ occupancy floor and the
$\delta = 0.5$ parsimony tolerance, are fixed \emph{a priori} on
interpretability and operational grounds rather than tuned. Their
sensitivity to alternative values, and the stability of $K^*$ under
resampling, remain untested and are left as follow-up work. Finally,
neither the corrected score nor filtered inference recovers the
$42.1\%$ of fraud dispersed through the manifold's body
(\cref{sec:latent-manifold}), as those transactions are, in the encoder's
own representation, indistinguishable from legitimate activity,
and no scoring-formula correction changes what the representation itself
has or has not captured.

\section{Conclusion}
\label{sec:conclusion}

We presented a three-tiered progression of Hidden Markov Models for
sequential fraud detection, culminating in a Neural VBEM HMM that
replaces a fixed emission family with a learned encoder while retaining
exact Bayesian forward-backward inference. A UMAP projection of its
latent space showed that the discovered regimes are
ordered segments of a single continuous behavioural manifold, with
confirmed fraud concentrated at its extreme, a finding that simultaneously
explains the model's residual AUPRC gap against a discriminative baseline
and confirms the sequential regime hypothesis this work is built on. We
showed that the model's natural output, regime-membership probability, is
routinely conflated with fraud probability, quantified the resulting
miscalibration as the Mean Regime Interpretation Error, and gave a
corrected score that is exact in expectation by construction. We further
separated batch inference, which uses look-ahead unavailable at
deployment time, from filtered inference, and reported both.

Taken together, these results argue against
treating this model as a drop-in replacement for a tuned discriminative
ranker, as it trails XGBoost on raw discrimination for reasons that are
structural as opposed to fixable by further tuning. What it offers instead
is what a transaction-level classifier cannot: a calibrated, interpretable
belief state over a customer's behavioural trajectory, updated at constant
cost per transaction, best used as a triage and interpretability layer
alongside a discriminative ranker rather than in competition with it.

\section*{Data and Code Availability}
The IEEE-CIS data are distributed through the original competition
portal \citep{ieeeCIS2019}, subject to its access terms. The transaction
data are not redistributed with this manuscript. A public implementation
and complete experiment manifest are not supplied with this version;
independent reproduction therefore remains limited.

\bibliography{references}

\newpage

\appendix

\section{Notation}
\label{app:notation}

The notation used throughout this work draws on several foundations in
sequential modelling and variational
methods, documented at the end of each table below. One convention
maintained throughout is $\mathbb{P}(\cdot)$ which denotes a probability measure
over a discrete event (a state assignment or a transition), while $p(\cdot)$
denotes a density evaluated at a point over a continuous variable (an
emission likelihood, a parameter prior), following Bishop
\citep{bishopPatternRecognitionMachine2007} and Rabiner
\citep{rabinerTutorialHiddenMarkov1989}.

\begingroup
\setstretch{1.3}
\setlength{\aboverulesep}{0.7ex}
\setlength{\belowrulesep}{0.7ex}

\begin{longtable}{
        >{\arraybackslash}p{2.6cm}
        >{\arraybackslash}p{3.4cm}
        >{\raggedright\arraybackslash}p{9cm}
    }
    \caption{Dataset and Sequence Notation} \label{tab:dataset-notation}                                                                                                                                                   \\
    \toprule
    \textbf{Symbol}                                                       & \textbf{Form}    & \textbf{Description}                                                                                                        \\
    \midrule
    \endfirsthead

    \multicolumn{3}{c}{\vspace{10pt}\textbf{Table \thetable{}}: continued from previous page\vspace{0.1pt}}                                                                                                                \\
    \toprule
    \textbf{Symbol}                                                       & \textbf{Form}    & \textbf{Description}                                                                                                        \\
    \midrule
    \endhead

    \midrule
    \multicolumn{3}{r}{\textit{Continued on next page}}                                                                                                                                                                    \\
    \endfoot

    \bottomrule
    \endlastfoot

    $\mathcal{D}$                                                         & Set              & Full dataset of $N$ labelled transactions.                                                                                  \\
    $N$                                                                   & Scalar integer   & Total number of transactions.                                                                                               \\
    $U$                                                                   & Scalar integer   & Total number of unique customers.                                                                                           \\
    $u$                                                                   & Index            & Customer index, $u = 1,\ldots,U$.                                                                                           \\
    $n$                                                                   & Index            & Global transaction index, $n = 1,\ldots,N$.                                                                                 \\
    $t$                                                                   & Index            & Chronological index within customer $u$'s sequence, $t=1,\ldots,T_u$.                                                       \\
    $\tau_{u,t}$                                                          & Scalar           & Transaction timestamp.                                                                                                      \\
    $T_u$                                                                 & Scalar integer   & Length of customer $u$'s sequence.                                                                                          \\
    $T_{\min}$                                                            & Scalar integer   & Minimum sequence length for direct fitting; below this, the hierarchical prior extension applies (\cref{sec:hierarchical}). \\
    $y_n$                                                                 & Binary           & Ground-truth fraud label, 1 = fraud. Used only for evaluation.                                                              \\
    $\mathbf{x}_n$, $\mathbf{X}_u$                                        & Vector, Sequence & Feature vector for transaction $n$; the ordered sequence for customer $u$.                                                  \\
    $\mathbf{x}^{\mathrm{cont}}_{u,t}$, $\mathbf{x}^{\mathrm{cat}}_{u,t}$ & Vectors          & Continuous and categorical feature blocks of transaction $(u,t)$.                                                           \\
    $F$, $D$, $J$                                                         & Scalar integers  & Total, continuous, and categorical feature counts ($F=D+J$).                                                                \\
    $C_j$                                                                 & Scalar integer   & Number of categories for categorical feature $j$.                                                                           \\
\end{longtable}
\endgroup

\begingroup
\setstretch{1.3}
\setlength{\aboverulesep}{0.7ex}
\setlength{\belowrulesep}{0.7ex}

\begin{longtable}{
        >{\arraybackslash}p{2.6cm}
        >{\arraybackslash}p{3.4cm}
        >{\raggedright\arraybackslash}p{9cm}
    }
    \caption{Hidden Markov Model Notation} \label{tab:hmm-notation}                                                                                                                                                    \\
    \toprule
    \textbf{Symbol}                                                                            & \textbf{Form}          & \textbf{Description}                                                                         \\
    \midrule
    \endfirsthead

    \multicolumn{3}{c}{\vspace{10pt}\normalsize{\textbf{Table \thetable{}}}: continued from previous page\vspace{0.1pt}}                                                                                               \\
    \toprule
    \textbf{Symbol}                                                                            & \textbf{Form}          & \textbf{Description}                                                                         \\
    \midrule
    \endhead

    \midrule
    \multicolumn{3}{r}{\textit{Continued on next page}}                                                                                                                                                                \\
    \endfoot

    \bottomrule
    \endlastfoot
    $K$, $K^*$                                                                                 & Scalar integers        & Candidate and selected model order.                                                          \\
    $k,i,j$                                                                                    & State indices          & General, from-, and to-state indices over $1,\ldots,K$.                                      \\
    $q_{u,t}$, $\mathbf{q}_u$                                                                  & Latent, Latent path    & Hidden state at $t$; full hidden state sequence.                                             \\
    $k^*$                                                                                      & Scalar integer         & Index of the fraud-associated state, identified post-training as $\arg\max_k \eta_k$.        \\
    $\bm\theta$                                                                                & Parameter set          & All HMM parameters: $\bm\pi, A, \bm\mu_k, \bm\Sigma_k, \bm\pi_k^{(j)}$.                      \\
    $\bm\pi$, $\pi_k$                                                                          & Vector, scalar         & Initial state distribution; probability of starting in state $k$.                            \\
    $A$, $A_{ij}$, $A_{kk}$                                                                    & Matrix, scalars        & Transition matrix; entry $i\to j$; self-transition (stickiness) of state $k$.                \\
    $B_k(\mathbf{x}_{u,t})$, $\mathbf{b}_t$                                                    & Scalar, vector         & Emission probability of $\mathbf{x}_{u,t}$ under state $k$; emission vector at $t$.          \\
    $\bm\mu_k$, $\bm\Sigma_k$, $\sigma^2_{k,d}$                                                & Vector, matrix, scalar & Gaussian emission mean, covariance, and its $d$-th diagonal variance for state $k$.          \\
    $\pi^{(j)}_{k,c}$, $\bm\pi^{(j)}_k$                                                        & Scalar, vector         & Categorical emission probability of category $c$; full vector over feature $j$'s categories. \\
    $\hat\pi_k$, $\hat A_{i,j}$, $\hat{\bm\mu}_k$, $\hat\sigma^2_{k,d}$, $\hat\pi^{(j)}_{k,c}$ & —                      & Baum-Welch (maximum-likelihood) parameter updates.                                           \\
\end{longtable}
\endgroup

Murphy \citep{murphyMachineLearningProbabilistic2012} presents
HMMs under the same $K$-state convention used here and serves as a
secondary standardising reference alongside Bishop
\citep{bishopPatternRecognitionMachine2007} for the departure from
Rabiner's \citep{rabinerTutorialHiddenMarkov1989}
$N$-state, $M$-symbol notation. Readers familiar with the latter should
treat $K$ here as Rabiner's $N$. The
per-customer subscript $u$ and the multi-sequence panel structure indexed
by $u=1,\ldots,U$ are extensions for transaction data and not attributable to
a single prior source. The EM framework is due to Dempster et al.\
\citep{dempsterMaximumLikelihoodIncomplete1977}; the Baum-Welch
re-estimation algorithm to Baum et al.\
\citep{baumMaximizationTechniqueOccurring1970},
building on Baum and Petrie \citep{baumStatisticalInferenceProbabilistic1966}; BIC model
selection to Schwarz \citep{schwarzEstimatingDimensionModel1978}.

\begingroup
\setstretch{1.3}
\setlength{\aboverulesep}{0.7ex}
\setlength{\belowrulesep}{0.7ex}

\begin{longtable}{
        >{\arraybackslash}p{2.6cm}
        >{\arraybackslash}p{3.4cm}
        >{\raggedright\arraybackslash}p{9cm}
    }
    \caption{Forward-Backward Variables} \label{tab:fb-notation}                                                                                                   \\
    \toprule
    \textbf{Symbol}               & \textbf{Form}  & \textbf{Description}                                                                                          \\
    \midrule
    \endfirsthead

    \multicolumn{3}{c}{\vspace{10pt}\normalsize{\textbf{Table \thetable{}}}: continued from previous page\vspace{0.1pt}}                                           \\
    \toprule
    \textbf{Symbol}               & \textbf{Form}  & \textbf{Description}                                                                                          \\
    \midrule
    \endhead

    \midrule
    \multicolumn{3}{r}{\textit{Continued on next page}}                                                                                                            \\
    \endfoot

    \bottomrule
    \endlastfoot

    $\alpha_t(k)$, $\bm\alpha$    & Scalar, matrix & Forward variable; full forward matrix for a sequence.                                                         \\
    $\beta_t(k)$, $\bm\beta$      & Scalar, matrix & Backward variable; full backward matrix for a sequence.                                                       \\
    $\gamma_t(k)$, $\bm\gamma$    & Scalar, matrix & Batch (smoothed) state-occupancy posterior; full matrix for a sequence.                                       \\
    $\gamma^{\mathrm{filt}}_t(k)$ & Scalar         & Filtered (forward-only) state-occupancy posterior: conditions on $\mathbf{x}_{u,1:t}$ only, no backward pass. \\
    $\xi_t(i,j)$, $\bm\xi$        & Scalar, tensor & Transition posterior; full tensor for a sequence.                                                             \\
    $\odot$                       & Operator       & Element-wise multiplication.                                                                                  \\
    $a^*$                         & Scalar         & Log-sum-exp stabilisation constant.                                                                           \\
    $\mathbf{1}[\cdot]$           & Operator       & Indicator function.                                                                                           \\
\end{longtable}
\endgroup

The forward, backward, state-occupancy, and transition-occupancy
posteriors $\alpha_t(k)$, $\beta_t(k)$, $\gamma_t(k)$, $\xi_t(i,j)$ follow
the conventions of Rabiner \citep{rabinerTutorialHiddenMarkov1989}.
$\gamma^{\mathrm{filt}}_t(k)$ is this work's own notation, introduced to
distinguish filtered from smoothed inference (\cref{sec:inference-scoring}).

\begingroup
\setstretch{1.3}
\setlength{\aboverulesep}{0.7ex}
\setlength{\belowrulesep}{0.7ex}

\begin{longtable}{
        >{\arraybackslash}p{2.6cm}
        >{\arraybackslash}p{3.4cm}
        >{\raggedright\arraybackslash}p{9cm}
    }
    \caption{Variational Bayesian Inference Notation} \label{tab:vbem-notation}                                                                                                                              \\
    \toprule
    \textbf{Symbol}                                                              & \textbf{Form}             & \textbf{Description}                                                                          \\
    \midrule
    \endfirsthead

    \multicolumn{3}{c}{\vspace{10pt}\normalsize{\textbf{Table \thetable{}}}: continued from previous page\vspace{0.1pt}}                                                                                     \\
    \toprule
    \textbf{Symbol}                                                              & \textbf{Form}             & \textbf{Description}                                                                          \\
    \midrule
    \endhead

    \midrule
    \multicolumn{3}{r}{\textit{Continued on next page}}                                                                                                                                                      \\
    \endfoot

    \bottomrule
    \endlastfoot

    $p(\mathbf{X}_u \mid \theta)$, $p(\mathbf{X}_u,\mathbf{q}_u \mid \theta)$    & Scalars                   & Marginal and joint sequence probabilities.                                                    \\
    $p(\bm\theta)$, $p(\bm\theta\mid\mathbf{X})$, $p(\mathbf{X})$                & Distributions, scalar     & Prior, true posterior, and model evidence.                                                    \\
    $q(\cdot)$, $q_i(\theta_i)$                                                  & Distributions             & Variational posterior; its mean-field factors.                                                \\
    $D_{\mathrm{KL}}(p\Vert q)$                                                  & Functional                & Kullback-Leibler divergence.                                                                  \\
    $\mathbb{E}_q[\cdot]$                                                        & Operator                  & Expectation under $q$.                                                                        \\
    $\mathcal{L}(q)$                                                             & Scalar                    & Evidence Lower Bound.                                                                         \\
    $\Gamma(\cdot)$, $\psi(\cdot)$                                               & Functions                 & Gamma function; digamma function.                                                             \\
    $\alpha_k$, $\alpha_{ij}$, $\alpha_{ii}$, $\kappa$                           & Hyperparameters           & Dirichlet concentrations for $\bm\pi$ and $A$; sticky self-transition entry and its strength. \\
    $\mathbf{m}_0,\kappa_0,\nu_0,\mathbf{S}_0$                                   & Hyperparameters           & NIW prior: mean, mean-confidence, degrees of freedom, scale matrix.                           \\
    $\alpha^{(j)}_0$                                                             & Hyperparameter            & Symmetric Dirichlet concentration for categorical feature $j$.                                \\
    $\tilde\alpha_k,\tilde\alpha_{ij},\tilde{\bm\alpha}_\pi,\tilde{\bm\alpha}_A$ & Posterior hyperparameters & Posterior Dirichlet parameters (scalar, then full vector/matrix forms).                       \\
    $\tilde{\mathbf{m}}_k,\tilde\kappa_k,\tilde\nu_k,\tilde{\mathbf{S}}_k$       & Posterior hyperparameters & Posterior NIW parameters for state $k$.                                                       \\
\end{longtable}
\endgroup

\begingroup
\setstretch{1.3}
\setlength{\aboverulesep}{0.7ex}
\setlength{\belowrulesep}{0.7ex}

\begin{longtable}{
        >{\arraybackslash}p{2.6cm}
        >{\arraybackslash}p{3.4cm}
        >{\raggedright\arraybackslash}p{9cm}
    }
    \caption{Neural VBEM Notation} \label{tab:neural-vbem-notation}                                                                                                                                                                                               \\
    \toprule
    \textbf{Symbol}                                                                          & \textbf{Form}        & \textbf{Description}                                                                                                                        \\
    \midrule
    \endfirsthead

    \multicolumn{3}{c}{\vspace{10pt}\normalsize{\textbf{Table \thetable{}}}: continued from previous page\vspace{0.1pt}}                                                                                                                                          \\
    \toprule
    \textbf{Symbol}                                                                          & \textbf{Form}        & \textbf{Description}                                                                                                                        \\
    \midrule
    \endhead

    \midrule
    \multicolumn{3}{r}{\textit{Continued on next page}}                                                                                                                                                                                                           \\
    \endfoot

    \bottomrule
    \endlastfoot

    $f_\theta$                                                                               & Function             & Neural encoder, $f_\theta:\mathbb{R}^F\to\mathbb{R}^{d_z}$.                                                                                 \\
    $\mathbf{z}_{u,t}$, $d_z$, $d_h$                                                         & Vector, integers     & Latent embedding; latent and hidden dimensions ($d_z\in\{32,64,128\}$, $d_h=512$).                                                          \\
    $d_{e_j}$, $E_j$                                                                         & Integer, matrix      & Embedding dimension and table for categorical feature $j$.                                                                                  \\
    $\hat{\mathbf{x}}$                                                                       & Vector               & Normalised continuous features fed to the encoder MLP (LayerNorm output).                                                                   \\
    $\mathrm{LN}(\cdot)$, $\mathrm{GELU}(\cdot)$                                             & Operator, activation & Layer Normalisation; Gaussian Error Linear Unit.                                                                                            \\
    $\mathbf{h}^{\mathrm{cont}}$, $\mathbf{h}^{\mathrm{cat}}$, $\mathbf{h}^{\mathrm{fused}}$ & Vectors              & Continuous-pathway, categorical-pathway, and fused hidden representations.                                                                  \\
    $W_1,b_1,W_2,b_2$                                                                        & Weights              & Continuous-feature MLP.                                                                                                                     \\
    $W_c,b_c$                                                                                & Weight, bias         & Categorical embedding projection.                                                                                                           \\
    $W_{\mathrm{fuse}},b_{\mathrm{fuse}},W_{\mathrm{proj}}$                                  & Weights              & Fusion block and final projection to $\mathbb{R}^{d_z}$.                                                                                    \\
    $\bm\mu^z_k$, $\bm\Sigma^z_k$                                                            & Vector, matrix       & Latent-space NIW posterior mean and covariance for state $k$.                                                                               \\
    $\tilde E_k$, $\tilde D_{u,t,k}$                                                         & Scalars              & Expected log-determinant of state $k$'s precision; expected squared Mahalanobis distance of $\mathbf{z}_{u,t}$ from $\tilde{\mathbf{m}}_k$. \\
    $\hat y_{u,t}$, $\omega$                                                                 & Scalar, scalar       & Phase-1 sigmoid classifier output; class weight ($\omega=N_-/N_+\approx27.6$).                                                              \\
    $N_+, N_-$                                                                               & Integers             & Fraud- and legitimate-labelled transaction counts.                                                                                          \\
    $w,b$                                                                                    & Weight, bias         & Temporary Phase-1 classification head, discarded after pretraining.                                                                         \\
    $\mathcal{L}(\theta,w,b)$                                                                & Scalar               & Phase-1 supervised pretraining loss.                                                                                                        \\
    $\mu_+,\mu_-,\Delta_\mu$                                                                 & Vectors, scalar      & Fraud/legitimate latent centroids; their separation, $\Delta_\mu=\Vert\mu_+-\mu_-\Vert_2$.                                                  \\
    $Z_u$                                                                                    & Matrix               & Frozen latent embeddings for customer $u$'s sequence, $Z_u\in\mathbb{R}^{T_u\times d_z}$.                                                   \\
    $\Vert\nabla\Vert$                                                                       & Scalar               & Gradient norm (clipped at 1.0).                                                                                                             \\
\end{longtable}
\endgroup

\begingroup
\setstretch{1.3}
\setlength{\aboverulesep}{0.7ex}
\setlength{\belowrulesep}{0.7ex}

\begin{longtable}{
        >{\arraybackslash}p{2.6cm}
        >{\arraybackslash}p{3.4cm}
        >{\raggedright\arraybackslash}p{9cm}
    }
    \caption{Hierarchical Prior Extension} \label{tab:hierarchical-notation}                                                                             \\
    \toprule
    \textbf{Symbol}                                         & \textbf{Form}  & \textbf{Description}                                                      \\
    \midrule
    \endfirsthead

    \multicolumn{3}{c}{\vspace{10pt}\normalsize{\textbf{Table \thetable{}}}: continued from previous page\vspace{0.1pt}}                                 \\
    \toprule
    \textbf{Symbol}                                         & \textbf{Form}  & \textbf{Description}                                                      \\
    \midrule
    \endhead

    \midrule
    \multicolumn{3}{r}{\textit{Continued on next page}}                                                                                                  \\
    \endfoot

    \bottomrule
    \endlastfoot

    $\mathbf{m}_0^{\mathrm{p}},\mathbf{S}_0^{\mathrm{p}}$   & Vector, matrix & Population-level NIW prior (moment-matched).                              \\
    $\hat{\mathbf{m}}_k^{(u)},\hat{\mathbf{S}}_k^{(u)}$     & Vector, matrix & Customer-$u$ converged posterior statistics feeding the population prior. \\
    $\mathcal{F}(\mathbf{x}_{u',t}\mid\mathbf{X}_{u',1:t})$ & Scalar         & Cold-start fraud score for new customer $u'$.                             \\
\end{longtable}
\endgroup

\begingroup
\setstretch{1.3}
\setlength{\aboverulesep}{0.7ex}
\setlength{\belowrulesep}{0.7ex}

\begin{longtable}{
        >{\arraybackslash}p{2.6cm}
        >{\arraybackslash}p{3.4cm}
        >{\raggedright\arraybackslash}p{9cm}
    }
    \caption{State Diagnostics and Model Selection} \label{tab:state-diagnostics-notation}                                                   \\
    \toprule
    \textbf{Symbol}                   & \textbf{Form}   & \textbf{Description}                                                               \\
    \midrule
    \endfirsthead

    \multicolumn{3}{c}{\vspace{10pt}\normalsize{\textbf{Table \thetable{}}}: continued from previous page\vspace{0.1pt}}                     \\
    \toprule
    \textbf{Symbol}                   & \textbf{Form}   & \textbf{Description}                                                               \\
    \midrule
    \endhead

    \midrule
    \multicolumn{3}{r}{\textit{Continued on next page}}                                                                                      \\
    \endfoot

    \bottomrule
    \endlastfoot

    $\mathcal{K}$, $\mathcal{K}^*$    & Sets            & Candidate model orders $\{2,\ldots,10\}$; the eligible subset.                     \\
    $N^{(K)}_k$, $\mathrm{o}^{(K)}_k$ & Scalars         & Effective count and fractional occupancy of state $k$ under order $K$.             \\
    $F^{(K)}_k$, $\eta^{(K)}_k$       & Scalars         & Fraud-weighted effective count; empirical fraud rate of state $k$ under order $K$. \\
    $k^{*(K)}$, $\mathrm{e}^{(K)}_k$  & Integer, scalar & Fraud-associated state under order $K$; its enrichment ratio.                      \\
    $\bar\eta$, $\delta$              & Scalars         & Population fraud rate ($\approx 3.5\%$); parsimony tolerance ($=0.5$).             \\
    $\mathrm{f}$                      & Notation        & Shorthand for ``fraud''.                                                           \\
\end{longtable}
\endgroup

\begingroup
\setstretch{1.3}
\setlength{\aboverulesep}{0.7ex}
\setlength{\belowrulesep}{0.7ex}

\begin{longtable}{
        >{\arraybackslash}p{2.6cm}
        >{\arraybackslash}p{3.4cm}
        >{\raggedright\arraybackslash}p{9cm}
    }
    \caption{Inference and Fraud Scoring} \label{tab:inference-notation}                                                                                                                                                                                \\
    \toprule
    \textbf{Symbol}                                      & \textbf{Form}          & \textbf{Description}                                                                                                                                                \\
    \midrule
    \endfirsthead

    \multicolumn{3}{c}{\vspace{10pt}\normalsize{\textbf{Table \thetable{}}}: continued from previous page\vspace{0.1pt}}                                                                                                                                \\
    \toprule
    \textbf{Symbol}                                      & \textbf{Form}          & \textbf{Description}                                                                                                                                                \\
    \midrule
    \endhead

    \midrule
    \multicolumn{3}{r}{\textit{Continued on next page}}                                                                                                                                                                                                 \\
    \endfoot

    \bottomrule
    \endlastfoot

    $\mathbf{c}_{u,t}$                                   & Vector, $\Delta^{K-1}$ & Transaction consequence: the real-time belief state, updated one transaction at a time from $\mathbf{c}_{u,t-1}$; coincides with $\gamma^{\mathrm{filt}}_t(\cdot)$. \\
    $\mathcal{F}_{\mathrm{batch}}(\mathbf{x}_{u,t})$     & Scalar                 & Batch fraud score, $\gamma_t(k^*)$.                                                                                                                                 \\
    $\mathcal{F}_{\mathrm{real}}(\mathbf{x}_{u,t})$      & Scalar                 & Real-time fraud score, $\mathbf{c}_{u,t}(k^*)$.                                                                                                                     \\
    $\mathcal{F}_{\mathrm{corrected}}(\mathbf{x}_{u,t})$ & Scalar                 & Posterior-predictive corrected fraud score, $\sum_k\gamma_t(k)\eta_k$.                                                                                              \\
    $\mathrm{MRIE}$                                      & Scalar                 & Mean Regime Interpretation Error, $\mathrm{o}_{k^*}(1-\eta_{k^*})$.                                                                                                 \\
\end{longtable}
\endgroup

The empirical per-state fraud rate $\eta^{(K)}_k$, the fraud-associated
state $k^*$, the population fraud rate $\bar\eta$, the transaction
consequence $\mathbf{c}_{u,t}$, and MRIE are this work's own notation and
are not attributable to a prior foundational source. The modelling
paradigm, applying an HMM trained on cardholder behaviour to flag
anomalously-placed posterior mass, follows Srivastava et al.\
\citep{srivastavaCreditCardFraud2008}.

\section{Derivations}
\label{app:derivations}

\subsection{The Evidence Lower Bound}

The starting point is the identity $p(\mathbf{X}) = p(\mathbf{X},\bm\theta)
    / p(\bm\theta\mid\mathbf{X})$, with Bayes' rule rearranged, the evidence is
the joint divided by the true posterior. Taking logs and an expectation
under the variational distribution $q(\bm\theta)$ on both sides (the
left-hand side does not involve $\bm\theta$, so this expectation leaves
it unchanged) gives

\begin{equation}
    \log p(\mathbf{X}) = \mathbb{E}_q\bigl[\log p(\mathbf{X},\bm\theta) - \log p(\bm\theta\mid\mathbf{X})\bigr].
\end{equation}

The right-hand side still contains
$p(\bm\theta\mid\mathbf{X})$, the intractable true posterior we are
trying to avoid computing directly. The next step introduces
$q(\bm\theta)$ into the expression itself, by adding and subtracting
$\log q(\bm\theta)$, creating the terms needed to isolate
$q$ from $p(\bm\theta\mid\mathbf{X})$:

\begin{equation}
    \log p(\mathbf{X}) = \mathbb{E}_q\bigl[\log p(\mathbf{X},\bm\theta) - \log q(\bm\theta) + \log q(\bm\theta) - \log p(\bm\theta\mid\mathbf{X})\bigr].
\end{equation}

Regrouping these four terms into two pairs, each pair becomes a ratio
inside its own expectation:

\begin{equation}
    \log p(\mathbf{X}) = \mathbb{E}_q\!\left[\log\frac{p(\mathbf{X},\bm\theta)}{q(\bm\theta)}\right]_{\mathcal{L}(q)}
    + \mathbb{E}_q\!\left[\log\frac{q(\bm\theta)}{p(\bm\theta\mid\mathbf{X})}\right]_{D_{\mathrm{KL}}(q\Vert p)}.
\end{equation}

The first term is the ELBO, $\mathcal{L}(q)$, where it depends only on $q$ and
the joint $p(\mathbf{X},\bm\theta)$, both computable. The second is
the KL divergence between $q(\bm\theta)$ and the true posterior,
the quantity we want, but cannot evaluate directly,
since it too depends on $p(\bm\theta\mid\mathbf{X})$.

Since $D_{\mathrm{KL}}(q\Vert p) \geq 0$ and $\log p(\mathbf{X})$ does not
depend on $q$, $\mathcal{L}(q) \leq \log p(\mathbf{X})$ always, maximising
$\mathcal{L}(q)$ is therefore equivalent to minimising
$D_{\mathrm{KL}}(q\Vert p)$, without ever having to evaluate it.

Under the mean-field factorisation (\cref{sec:vb-inference}),
$q(\bm\theta) = q(\bm\pi)\prod_i q(A_{i,:}) \prod_k q(\bm\mu_k,\bm\Sigma_k)
    \prod_{k,j} q(\bm\pi^{(j)}_k)$, and assuming the prior factorises
identically (each block was given its own independent prior in
\cref{sec:vb-inference}), the KL term in $\mathcal{L}(q)$ decomposes into
one closed-form divergence per block, since the KL divergence between two
distributions that factorise the same way over the same partition is the
sum of the per-factor KL divergences. Each block contributes its own
term, defined as follows.

\begin{equation}
    D^{\bm\pi}_{\mathrm{KL}} = D_{\mathrm{KL}}\bigl(q(\bm\pi)\Vert p(\bm\pi)\bigr).
\end{equation}

The divergence of the initial-state posterior from its Dirichlet prior.

\begin{equation}
    D^{A}_{\mathrm{KL}} = \sum_{i=1}^K D_{\mathrm{KL}}\bigl(q(A_{i,:})\Vert p(A_{i,:})\bigr).
\end{equation}

The same divergence summed across every row of the transition matrix,
one term per originating state $i$.

\begin{equation}
    D^{\mu\Sigma}_{\mathrm{KL}} = \sum_{k=1}^K D_{\mathrm{KL}}\bigl(q(\bm\mu_k,\bm\Sigma_k)\Vert p(\bm\mu_k,\bm\Sigma_k)\bigr).
\end{equation}

The summed divergence of each state's Gaussian emission parameters from
its NIW prior, that is, the term that regularises the fraud state specifically,
since it is the one with the least data behind it.

\begin{equation}
    D^{\pi^{(j)}}_{\mathrm{KL}} = \sum_{k=1}^K\sum_{j=1}^J D_{\mathrm{KL}}\bigl(q(\bm\pi^{(j)}_k)\Vert p(\bm\pi^{(j)}_k)\bigr).
\end{equation}

The equivalent sum across every categorical emission distribution, one
term per state-feature pair.

With all four terms defined, the ELBO itself is

\begin{equation}
    \mathcal{L}(q) = \mathbb{E}_q\bigl[\log p(\mathbf{X}\mid\bm\theta)\bigr] - D^{\bm\pi}_{\mathrm{KL}} - D^{A}_{\mathrm{KL}} - D^{\mu\Sigma}_{\mathrm{KL}} - D^{\pi^{(j)}}_{\mathrm{KL}}.
    \label{eq:elbo-expanded}
\end{equation}

A state that receives little support from the data keeps a posterior
close to its prior in every one of these terms, so its contribution to
the total penalty stays small automatically, forming the data-dependent
complexity control that a fixed-penalty criterion like BIC
(\cref{sec:vb-inference}) does not have.

Each Dirichlet-Dirichlet term above has the standard closed form: for
$q=\mathrm{Dir}(\tilde{\bm\alpha})$ against $p=\mathrm{Dir}(\bm\alpha)$
over $K$ categories, writing $B(\cdot)$ for the multivariate Beta
function (a Dirichlet's normalising constant),

\begin{equation}
    \log\frac{B(\bm\alpha)}{B(\tilde{\bm\alpha})} = \log\Gamma\! \ \Bigl(\textstyle\sum_k\tilde\alpha_k\Bigr) - \sum_k\log\Gamma(\tilde\alpha_k) - \log\Gamma\! \ \Bigl(\textstyle\sum_k\alpha_k\Bigr) + \sum_k\log\Gamma(\alpha_k).
\end{equation}

The log-ratio of the two distributions' normalising constants, that is, how
much more (or less) probability mass the posterior's shape concentrates
relative to the prior's, independent of where that mass sits.

\begin{equation}
    W = \sum_k(\tilde\alpha_k-\alpha_k)\Bigl[\psi(\tilde\alpha_k)-\psi\bigl(\textstyle\sum_k\tilde\alpha_k\bigr)\Bigr].
\end{equation}

The only term that involves both distributions' parameters directly, as it
weights each category's concentration difference $(\tilde\alpha_k-\alpha_k)$
by the expected log-probability under $q$, so a category where the
posterior has moved far from the prior contributes more to the
divergence than one where it has barely moved.

\begin{equation}
    D_{\mathrm{KL}}\bigl(\mathrm{Dir}(\tilde{\bm\alpha})\Vert\mathrm{Dir}(\bm\alpha)\bigr) = \log\frac{B(\bm\alpha)}{B(\tilde{\bm\alpha})} + W,
\end{equation}

applied once for $q(\bm\pi)$ and once per row for each $q(A_{i,:})$. The
NIW-NIW divergence for $q(\bm\mu_k,\bm\Sigma_k)$ against its prior is
likewise closed-form under conjugacy \citep{bealVariationalAlgorithmsApproximate2003},
combining a Gaussian term in $(\tilde{\mathbf{m}}_k,\tilde\kappa_k)$ with a
Wishart term in $(\tilde\nu_k,\tilde{\mathbf{S}}_k)$ via trace and
log-determinant operations on the precision matrices; we omit the full
multi-line expression here since it is standard and not specific to this
work, and use \cref{eq:elbo-expanded} directly for model selection
(\cref{sec:model-selection}) via its computed value at convergence.

\subsection{Expected Log-Parameters}

The VBEM E-step (\cref{sec:vb-inference}) replaces each point-estimate
log-parameter with its expectation under the current Dirichlet posterior.
For $\pi\sim\mathrm{Dir}(\tilde{\bm\alpha}_\pi)$ and each row
$A_{i,:}\sim\mathrm{Dir}(\tilde{\bm\alpha}_{i,:})$, the standard Dirichlet
expected-log-parameter identity, via the digamma function $\psi(x)$, gives

\begin{equation}
    \mathbb{E}_q[\log\pi_k] = \psi(\tilde\alpha_k) - \psi\Bigl(\textstyle\sum_{k'}\tilde\alpha_{k'}\Bigr),
    \qquad
    \mathbb{E}_q[\log A_{ij}] = \psi(\tilde\alpha_{ij}) - \psi\Bigl(\textstyle\sum_{j'}\tilde\alpha_{ij'}\Bigr).
\end{equation}

These substitute directly for $\log\pi_k$ and $\log A_{ij}$ in the
forward-backward recursion (\cref{sec:hmm-framework}), unchanged.
The expected latent-space log-emission used by both the VBEM
and Neural VBEM tiers follows the same principle applied to the NIW
posterior:

\begin{equation}
    \mathbb{E}_q\bigl[\log p(\mathbf{z}_{u,t}\mid q_{u,t}=k)\bigr]
    = -\frac{d_z}{2}\log(2\pi) + \frac{1}{2}\tilde E_k - \frac{1}{2}\tilde D_{u,t,k} - \frac{d_z}{2\tilde\kappa_k},
\end{equation}

where

\begin{equation}
    \tilde E_k = \sum_{d=1}^{d_z}\bigl[\psi(\tilde\nu_k/2) -
        \log(\tilde S_{k,dd}/2)\bigr],
\end{equation}

is the expected log-determinant of state
$k$'s precision (summed over the diagonal, under the diagonal-covariance
restriction, \cref{sec:hmm-framework}), and

\begin{equation}
    \tilde D_{u,t,k} =
    \sum_{d=1}^{d_z} (z_{u,t,d}-\tilde m_{k,d})^2 \cdot (\tilde\nu_k /
    \tilde S_{k,dd}),
\end{equation}

is the expected squared Mahalanobis distance from
$\mathbf{z}_{u,t}$ to the posterior mean. The final term,
$d_z/(2\tilde\kappa_k)$, is the correction for uncertainty in
$\tilde{\mathbf{m}}_k$ itself, absent from the Baum-Welch point estimate
but present in both Bayesian tiers.

\subsection{Baum-Welch M-Step}

Given the posterior state-occupancy $\gamma_{u,t}(k)$ and
transition-occupancy $\xi_{u,t}(i,j)$ from the E-step, the closed-form
maximum-likelihood updates are as follows.

\begin{equation}
    \hat\pi_k = \frac{1}{U}\displaystyle\sum_{u=1}^U \gamma_{u,1}(k).
\end{equation}

The empirical frequency with which each state starts a sequence,
averaged across all $U$ customers.

\begin{equation}
    \hat A_{ij} = \frac{\displaystyle\sum_{u=1}^U \displaystyle\sum_{t=1}^{T_u-1} \xi_{u,t}(i,j)}{\displaystyle\sum_{u=1}^U \displaystyle\sum_{t=1}^{T_u-1} \gamma_{u,t}(i)}.
\end{equation}

The numerator counts, in expectation, how often the chain actually moved
from $i$ to $j$; the denominator counts how often it was in $i$ at all
with a transition still to make. The ratio is the empirical transition
frequency, and the denominator's sum stops at $T_u-1$ because no
transition originates at a sequence's final timestep.

\begin{equation}
    \hat{\bm\mu}_k = \frac{\displaystyle\sum_{u,t} \gamma_{u,t}(k)\,\mathbf{x}^{\mathrm{cont}}_{u,t}}{\displaystyle\sum_{u,t} \gamma_{u,t}(k)}.
\end{equation}

A responsibility-weighted average, where each transaction contributes to state
$k$'s mean in proportion to how likely it is to belong to $k$, rather
than by a hard assignment.

\begin{equation}
    \hat\sigma^2_{k,d} = \frac{\displaystyle\sum_{u,t} \gamma_{u,t}(k)\,\bigl(x^{\mathrm{cont}}_{u,t,d} - \hat\mu_{k,d}\bigr)^2}{\displaystyle\sum_{u,t} \gamma_{u,t}(k)}.
\end{equation}

The corresponding weighted variance, computed using the mean just
updated above, that is, how spread out feature $d$ is among the transactions
currently believed to belong to state $k$.

\begin{equation}
    \hat\pi^{(j)}_{k,c} = \frac{\displaystyle\sum_{u,t} \gamma_{u,t}(k)\,\mathbf{1}\bigl[x^{\mathrm{cat}}_{u,t,j}=c\bigr]}{\displaystyle\sum_{u,t} \gamma_{u,t}(k)}.
\end{equation}

The categorical analogue of the same responsibility-weighting idea, that is, the
weighted proportion of state $k$'s mass that lands on category $c$ of
feature $j$.

These are standard EM re-estimation formulae
\citep{rabinerTutorialHiddenMarkov1989,dempsterMaximumLikelihoodIncomplete1977}
adapted to the multi-sequence, mixed-emission setting of
\cref{sec:hmm-framework}.

\subsection{VBEM Posterior Updates}

With effective count $N_k = \sum_{u,t}\gamma_{u,t}(k)$, the Dirichlet
posteriors update as simple pseudo-count accumulation,

\begin{equation}
    \tilde\alpha_k = \alpha_k + \sum_{u=1}^U \gamma_{u,1}(k), \qquad
    \tilde\alpha_{ij} = \alpha_{ij} + \sum_{u=1}^U\sum_{t=1}^{T_u-1} \xi_{u,t}(i,j).
\end{equation}

Both are the prior concentration plus the expected count of the
corresponding event, as the posterior simply accumulates pseudo-counts on
top of the prior, which is the defining behaviour of Dirichlet-categorical
conjugacy and the reason these updates need no numerical optimisation.

The NIW posterior updates, with weighted sample mean $\hat{\bm\mu}_k =
    \bigl(\sum_{u,t}\gamma_{u,t}(k)\,\mathbf{x}^{\mathrm{cont}}_{u,t}\bigr)/N_k$
identical in form to the Baum-Welch mean update above, are as follows.

\begin{equation}
    \tilde\kappa_k = \kappa_0 + N_k.
\end{equation}

Posterior confidence in the mean grows one-for-one with the effective
evidence $N_k$, as a state with many assigned transactions has a large
$\tilde\kappa_k$, and its posterior mean becomes correspondingly harder
to move with further data.

\begin{equation}
    \tilde{\mathbf{m}}_k = \frac{\kappa_0\mathbf{m}_0 + N_k\hat{\bm\mu}_k}{\tilde\kappa_k}.
\end{equation}

A precision-weighted average of the prior mean and the data mean
$\hat{\bm\mu}_k$, as $N_k$ grows, $\tilde{\mathbf{m}}_k$ moves away from
$\mathbf{m}_0$ and toward the data, at a rate set by how the two
confidences, $\kappa_0$ and $N_k$, compare.

\begin{equation}
    \tilde\nu_k = \nu_0 + N_k.
\end{equation}

Posterior degrees of freedom, controlling confidence in the covariance
shape, grow the same way as $\tilde\kappa_k$ does for the mean.

\begin{equation}
    \tilde{\mathbf{S}}_k = \mathbf{S}_0 + \sum_{u,t}\gamma_{u,t}(k)\bigl(\mathbf{x}^{\mathrm{cont}}_{u,t}-\hat{\bm\mu}_k\bigr)\bigl(\mathbf{x}^{\mathrm{cont}}_{u,t}-\hat{\bm\mu}_k\bigr)^\top + \frac{\kappa_0 N_k}{\tilde\kappa_k}\bigl(\hat{\bm\mu}_k-\mathbf{m}_0\bigr)\bigl(\hat{\bm\mu}_k-\mathbf{m}_0\bigr)^\top.
\end{equation}

The first added term is the within-state scatter of the data assigned to
state $k$; the second is a correction for the uncertainty introduced by
the sample mean's own distance from the prior mean, which expands the
posterior covariance exactly when the data disagrees with the prior.

For the Neural VBEM tier, these same four update equations apply
unchanged with $\mathbf{x}^{\mathrm{cont}}_{u,t}$ replaced by
$\mathbf{z}_{u,t}$ throughout (\cref{sec:neural-extension}).

\section{Extended Model Selection Results}
\label{app:extended-model-selection}

This section gives the full per-$K$ trail behind
\cref{tab:model-selection-summary}: model-complexity curves, complete
occupancy/fraud-rate/enrichment tables, and transition-matrix structures.

\subsection{Model Complexity: BIC vs ELBO}

Before any state-level diagnostic is examined, the raw model-order
selection curves are assessed: BIC for the Baum-Welch
tier, ELBO for VBEM, both swept across every candidate $K$ and both
feature sets, since the shape of these curves is
the first signal of how much structure
each model order actually recovers before the eligibility criteria of
\cref{sec:model-selection} are applied.

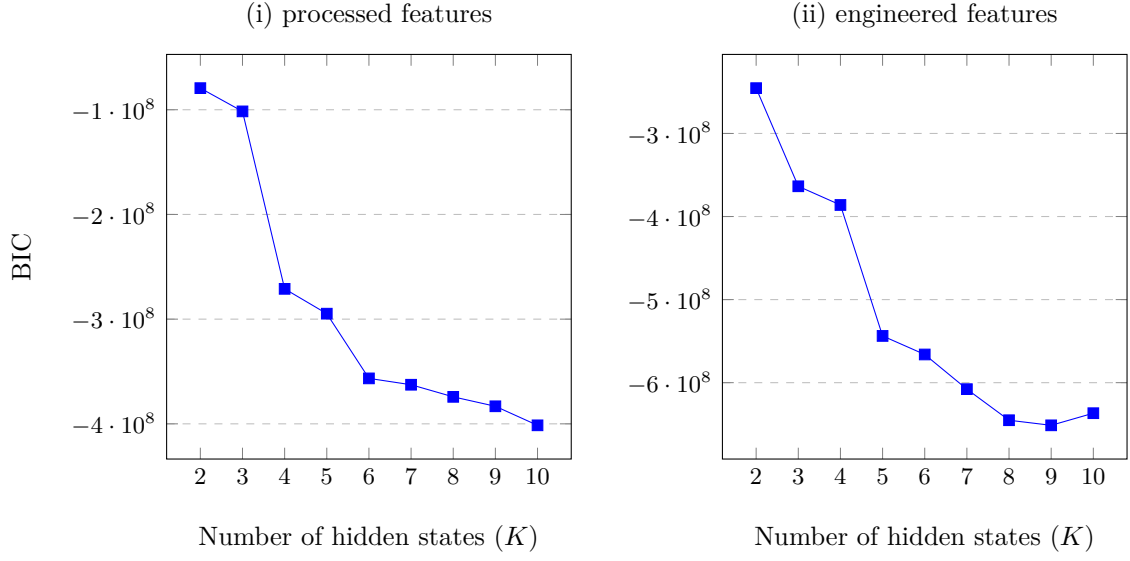
\begin{figure}[H]
    \centering
    \begin{tikzpicture}
        \begin{groupplot}[
                group style={group size=2 by 1, horizontal sep=2cm},
                width=0.42\textwidth, height=0.42\textwidth,
                xlabel={Number of hidden states ($K$)}, ylabel={BIC},
                xlabel shift=8pt, ylabel shift=8pt, xtick=data, enlargelimits=0.1,
                ymajorgrids=true, grid style=dashed, scaled y ticks=false,
                xlabel style={font=\small}, ylabel style={font=\small},
                title style={font=\small}, tick label style={font=\footnotesize},
            ]
            \nextgroupplot[title={(i) processed features}]
            \addplot[color=blue, mark=square*] coordinates {
                    (2,-79374808.35)(3,-101564387.73)(4,-271065667.19)(5,-294780324.73)
                    (6,-356586524.12)(7,-362655726.59)(8,-374218543.23)(9,-383235379.93)(10,-401410412.52)
                };
            \nextgroupplot[title={(ii) engineered features}, ylabel={}]
            \addplot[color=blue, mark=square*] coordinates {
                    (2,-245454908.40)(3,-363591305.07)(4,-386108338.66)(5,-543745690.91)
                    (6,-566063170.55)(7,-607807516.97)(8,-645212946.97)(9,-651338133.68)(10,-636753044.40)
                };
        \end{groupplot}
    \end{tikzpicture}
    \caption{Bayesian Information Criterion for the Baum-Welch HMM across
        $K\in\{2,\ldots,10\}$. Each panel's vertical axis is scaled to its own
        range; compare curve shape and minimum location within a panel, not
        absolute values across panels. Engineered features reach a minimum at
        $K=9$ before rising at $K=10$; processed features show no such elbow
        across the evaluated range.}
    \label{fig:appc-bw-bic}
\end{figure}

\begin{figure}[H]
    \centering
    \begin{tikzpicture}
        \begin{groupplot}[
                group style={group size=2 by 1, horizontal sep=2cm},
                width=0.42\textwidth, height=0.42\textwidth,
                xlabel={Number of hidden states ($K$)}, ylabel={ELBO},
                xlabel shift=8pt, ylabel shift=8pt, xtick=data, enlargelimits=0.1,
                ymajorgrids=true, grid style=dashed, scaled y ticks=false,
                xlabel style={font=\small}, ylabel style={font=\small},
                title style={font=\small}, tick label style={font=\footnotesize},
            ]
            \nextgroupplot[title={(i) processed features}]
            \addplot[color=red, mark=triangle*] coordinates {
                    (2,199658333.49)(3,286062276.43)(4,300258365.07)(5,304237815.51)
                    (6,329833885.88)(7,357945123.69)(8,354244967.05)(9,356291716.40)(10,364002397.17)
                };
            \nextgroupplot[title={(ii) engineered features}, ylabel={}]
            \addplot[color=red, mark=triangle*] coordinates {
                    (2,277935153.80)(3,333372461.07)(4,344339216.09)(5,373159632.21)
                    (6,372886930.77)(7,380373431.68)(8,375422033.68)(9,383062819.32)(10,474632121.02)
                };
        \end{groupplot}
    \end{tikzpicture}
    \caption{Evidence Lower Bound for the VBEM HMM across $K\in\{2,\ldots,10\}$.
        Unlike BIC, ELBO increases with $K$; the sharp rise at $K=10$ for
        engineered features ($+24\%$ over $K=9$) reflects a qualitatively
        different partition, examined further
        via the transition-matrix structures in \Cref{app:stickiness}}
    \label{fig:appc-vbem-elbo}
\end{figure}
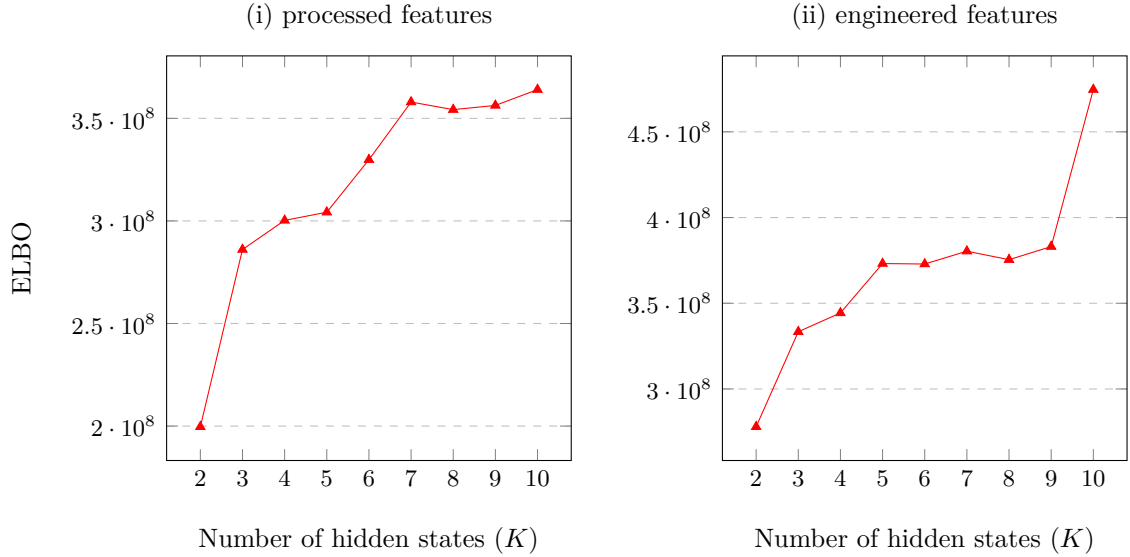

Favourable evidence alone does not guarantee an interpretable fraud
regime; the state-level diagnostics below, and the eligibility criteria
of \cref{sec:model-selection}, are what ultimately determine $K^*$.

\subsection{State Diagnostics: Baum-Welch HMM}
\label{sec:appc-bw-diagnostics}

\Cref{tab:appc-bw-batch,tab:appc-bw-filtered} give the full per-$K$
occupancy, fraud rate, and enrichment trail for the Baum-Welch tier
under both posterior modes and both feature sets, and a combined visual
comparison against VBEM across all three metrics follows at the end of
\cref{sec:appc-vbem-diagnostics} below.

\begin{table}[H]
    \centering
    \small
    \setstretch{1.3}
    \setlength{\aboverulesep}{0.7ex}
    \setlength{\belowrulesep}{0.7ex}
    \setlength{\tabcolsep}{3.2pt}
    \caption{Baum-Welch HMM: fraud-state occupancy, fraud rate, and
        enrichment under batch (smoothed) posteriors, $K\in\{2,\ldots,10\}$,
        processed (p) and engineered (e) features.}
    \label{tab:appc-bw-batch}
    \begin{tabular}{cS[table-format=1.3]S[table-format=1.3]S[table-format=1.3]S[table-format=1.3]S[table-format=1.3]S[table-format=1.3]}
        \toprule
        {\textbf{K}} & {\textbf{Occup. (p)}} & {\textbf{Occup. (e)}} & {\textbf{F. Rate (p)}} & {\textbf{F. Rate (e)}} & {\textbf{Enrich. (p)}} & {\textbf{Enrich. (e)}} \\
        \midrule
        2            & 0.242                 & 0.242                 & 0.082                  & 0.083                  & 2.337                  & 2.343                  \\
        3            & 0.219                 & 0.109                 & 0.090                  & 0.137                  & 2.536                  & 3.877                  \\
        4            & 0.073                 & 0.103                 & 0.162                  & 0.134                  & 4.596                  & 3.790                  \\
        5            & 0.041                 & 0.057                 & 0.163                  & 0.195                  & 4.618                  & 5.513                  \\
        6            & 0.045                 & 0.044                 & 0.216                  & 0.224                  & 6.125                  & 6.358                  \\
        7            & 0.040                 & 0.030                 & 0.219                  & 0.211                  & 6.194                  & 5.973                  \\
        8            & 0.043                 & \bfseries 0.018       & \bfseries 0.223        & 0.194                  & \bfseries 6.308        & 5.484                  \\
        9            & 0.046                 & 0.034                 & 0.222                  & \bfseries 0.268        & 6.300                  & \bfseries 7.598        \\
        10           & \bfseries 0.038       & 0.033                 & 0.200                  & 0.267                  & 5.651                  & 7.569                  \\
        \bottomrule
    \end{tabular}
\end{table}

\begin{table}[H]
    \centering
    \small
    \setstretch{1.3}
    \setlength{\aboverulesep}{0.7ex}
    \setlength{\belowrulesep}{0.7ex}
    \setlength{\tabcolsep}{3.2pt}
    \caption{Baum-Welch HMM: fraud-state occupancy, fraud rate, and
        enrichment under filtered (forward-only) posteriors, $K\in\{2,\ldots,10\}$,
        processed (p) and engineered (e) features.}
    \label{tab:appc-bw-filtered}
    \begin{tabular}{cS[table-format=1.3]S[table-format=1.3]S[table-format=1.3]S[table-format=1.3]S[table-format=1.3]S[table-format=1.3]}
        \toprule
        {\textbf{K}} & {\textbf{Occup. (p)}} & {\textbf{Occup. (e)}} & {\textbf{F. Rate (p)}} & {\textbf{F. Rate (e)}} & {\textbf{Enrich. (p)}} & {\textbf{Enrich. (e)}} \\
        \midrule
        2            & 0.242                 & 0.242                 & 0.082                  & 0.083                  & 2.337                  & 2.343                  \\
        3            & 0.219                 & 0.109                 & 0.090                  & 0.137                  & 2.536                  & 3.877                  \\
        4            & 0.073                 & 0.103                 & 0.162                  & 0.134                  & 4.596                  & 3.790                  \\
        5            & 0.041                 & 0.057                 & 0.163                  & 0.195                  & 4.618                  & 5.513                  \\
        6            & 0.045                 & 0.044                 & 0.216                  & 0.224                  & 6.125                  & 6.358                  \\
        7            & 0.040                 & 0.030                 & 0.219                  & 0.211                  & 6.194                  & 5.974                  \\
        8            & 0.043                 & \bfseries 0.018       & \bfseries 0.223        & 0.194                  & \bfseries 6.308        & 5.484                  \\
        9            & 0.046                 & 0.034                 & 0.222                  & \bfseries 0.268        & 6.300                  & \bfseries 7.598        \\
        10           & \bfseries 0.038       & 0.033                 & 0.200                  & 0.267                  & 5.652                  & 7.568                  \\
        \bottomrule
    \end{tabular}
\end{table}

Both feature sets agree closely at $K=2$ (occupancy $0.242$ both,
enrichment within $0.007\times$), consistent with a model this coarse
recovering little beyond the dominant binary partition regardless of
feature set. Enrichment for engineered features is non-monotonic between
$K=6$ and $K=9$ (peaking at $6.36\times$, dipping to $5.48\times$ at
$K=8$, then rising sharply to $7.60\times$ at $K=9$), consistent with
maximum-likelihood estimation growing less stable as evidence per state
thins with increasing $K$, as every occupancy value across both tables
comfortably clears the $0.5\%$ eligibility floor, the smallest being
$1.8\%$ at $K=8$, engineered.

\subsection{State Diagnostics: VBEM HMM}
\label{sec:appc-vbem-diagnostics}

\begin{table}[H]
    \centering
    \small
    \setstretch{1.3}
    \setlength{\aboverulesep}{0.7ex}
    \setlength{\belowrulesep}{0.7ex}
    \setlength{\tabcolsep}{3.2pt}
    \caption{VBEM HMM: fraud-state occupancy, fraud rate, and enrichment
        under batch (smoothed) posteriors, $K\in\{2,\ldots,10\}$, processed (p)
        and engineered (e) features.}
    \label{tab:appc-vbem-batch}
    \begin{tabular}{cS[table-format=1.3]S[table-format=1.3]S[table-format=1.3]S[table-format=1.3]S[table-format=1.3]S[table-format=1.3]}
        \toprule
        {\textbf{K}} & {\textbf{Occup. (p)}} & {\textbf{Occup. (e)}} & {\textbf{F. Rate (p)}} & {\textbf{F. Rate (e)}} & {\textbf{Enrich. (p)}} & {\textbf{Enrich. (e)}} \\
        \midrule
        2            & 0.242                 & 0.242                 & 0.082                  & 0.083                  & 2.337                  & 2.344                  \\
        3            & 0.085                 & 0.110                 & 0.135                  & 0.136                  & 3.824                  & 3.858                  \\
        4            & 0.071                 & 0.104                 & 0.161                  & 0.134                  & 4.549                  & 3.784                  \\
        5            & 0.050                 & 0.052                 & 0.134                  & 0.213                  & 3.809                  & 6.043                  \\
        6            & 0.026                 & 0.046                 & 0.152                  & 0.220                  & 4.311                  & 6.242                  \\
        7            & 0.056                 & 0.038                 & 0.186                  & 0.253                  & 5.261                  & 7.158                  \\
        8            & 0.042                 & 0.032                 & \textbf{0.239}         & 0.200                  & \textbf{6.780}         & 5.652                  \\
        9            & 0.048                 & \textbf{0.026}        & 0.217                  & \textbf{0.277}         & 6.156                  & \textbf{7.845}         \\
        10           & \textbf{0.022}        & 0.032                 & 0.233                  & 0.275                  & 6.599                  & 7.800                  \\
        \bottomrule
    \end{tabular}
\end{table}

\begin{table}[H]
    \centering
    \small
    \setstretch{1.3}
    \setlength{\aboverulesep}{0.7ex}
    \setlength{\belowrulesep}{0.7ex}
    \setlength{\tabcolsep}{3.2pt}
    \caption{VBEM HMM: fraud-state occupancy, fraud rate, and enrichment
        under filtered (forward-only) posteriors, $K\in\{2,\ldots,10\}$,
        processed (p) and engineered (e) features.}
    \label{tab:appc-vbem-filtered}
    \begin{tabular}{cS[table-format=1.3]S[table-format=1.3]S[table-format=1.3]S[table-format=1.3]S[table-format=1.3]S[table-format=1.3]}
        \toprule
        {\textbf{K}} & {\textbf{Occup. (p)}} & {\textbf{Occup. (e)}} & {\textbf{F. Rate (p)}} & {\textbf{F. Rate (e)}} & {\textbf{Enrich. (p)}} & {\textbf{Enrich. (e)}} \\
        \midrule
        2            & 0.242                 & 0.242                 & 0.082                  & 0.083                  & 2.337                  & 2.344                  \\
        3            & 0.085                 & 0.110                 & 0.135                  & 0.136                  & 3.824                  & 3.858                  \\
        4            & 0.071                 & 0.104                 & 0.161                  & 0.134                  & 4.549                  & 3.784                  \\
        5            & 0.050                 & 0.052                 & 0.134                  & 0.213                  & 3.809                  & 6.043                  \\
        6            & 0.026                 & 0.046                 & 0.152                  & 0.220                  & 4.311                  & 6.242                  \\
        7            & 0.056                 & 0.038                 & 0.186                  & 0.253                  & 5.261                  & 7.159                  \\
        8            & 0.042                 & 0.032                 & \textbf{0.239}         & 0.200                  & \textbf{6.779}         & 5.652                  \\
        9            & 0.048                 & \textbf{0.026}        & 0.217                  & \textbf{0.277}         & 6.156                  & \textbf{7.845}         \\
        10           & \textbf{0.022}        & 0.032                 & 0.233                  & 0.275                  & 6.599                  & 7.800                  \\
        \bottomrule
    \end{tabular}
\end{table}

At $K=2$, VBEM reproduces Baum-Welch almost exactly (occupancy $0.242$,
enrichment $2.34\times$ both), consistent with abundant per-state evidence
leaving the NIW prior no room to move the posterior away from the
maximum-likelihood estimate. The advantage of Bayesian regularisation
grows with $K$, as evidence per state thins: at $K^*=7$, engineered,
VBEM's enrichment ($7.16\times$) exceeds Baum-Welch's own peak
($6.36\times$ at $K=6$), and by $K=9$--$10$ VBEM leads Baum-Welch by
roughly $0.2$--$1.2\times$ at every matching order. Batch and filtered
numbers again agree to three decimals throughout.

Combined visual comparison against Baum-Welch follows for all three
diagnostics, batch (smoothed) posteriors in the top row of each figure,
filtered (forward-only) in the bottom row, so the near-agreement between
the two posterior modes reported numerically above is visible directly.

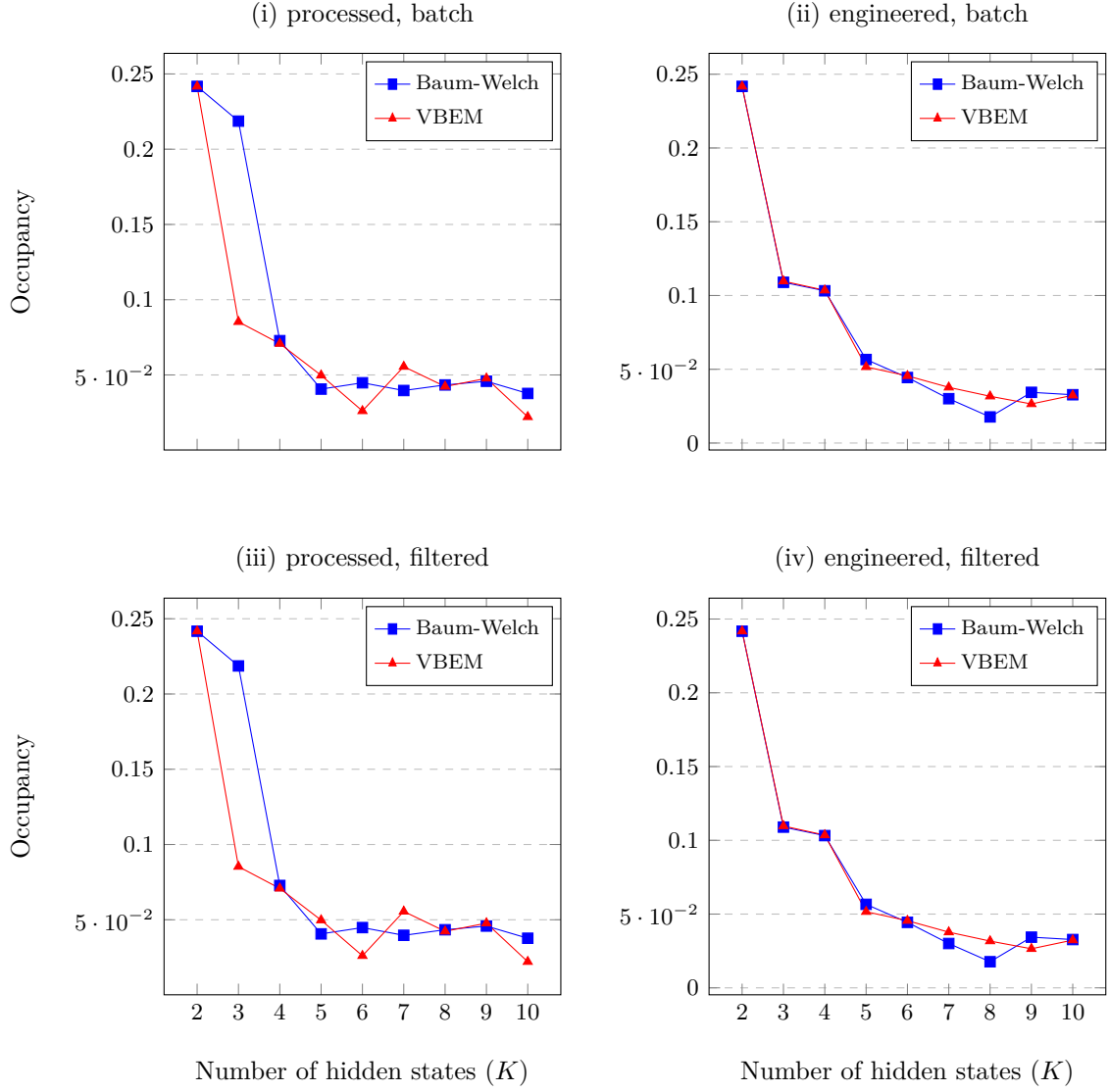
\begin{figure}[H]
    \centering
    \begin{tikzpicture}
        \begin{groupplot}[
                group style={group size=2 by 2, horizontal sep=2cm, vertical sep=2cm,
                        x descriptions at=edge bottom},
                width=0.42\textwidth, height=0.42\textwidth,
                xlabel={Number of hidden states ($K$)}, ylabel={Occupancy},
                xlabel shift=8pt, ylabel shift=8pt, xtick=data, enlargelimits=0.1,
                ymajorgrids=true, grid style=dashed, scaled y ticks=false,
                xlabel style={font=\small}, ylabel style={font=\small},
                title style={font=\small}, tick label style={font=\footnotesize},
                legend style={font=\scriptsize, inner sep=3pt, row sep=2pt, clip=false, at={(0.98,0.98)}, anchor=north east},
                legend cell align=left, legend image post style={xscale=0.8},
            ]
            \nextgroupplot[title={(i) processed, batch}, xlabel={}]
            \addplot[color=blue, mark=square*] coordinates {
                    (2,0.241716170011816)(3,0.21864568162595788)(4,0.07278199491772055)(5,0.040608411699771856)
                    (6,0.04479325395032762)(7,0.03965924578889485)(8,0.04335441386842013)(9,0.0458098389637972)(10,0.037701890669979754)
                };
            \addlegendentry{Baum-Welch}
            \addplot[color=red, mark=triangle*] coordinates {
                    (2,0.2417141432627822)(3,0.08531695118041477)(4,0.07081908322277529)(5,0.0497431284441549)
                    (6,0.026064943955449494)(7,0.05552660020029567)(8,0.04232211070800841)(9,0.04773164505285413)(10,0.02211242011582796)
                };
            \addlegendentry{VBEM}
            \nextgroupplot[title={(ii) engineered, batch}, ylabel={}, xlabel={}]
            \addplot[color=blue, mark=square*] coordinates {
                    (2,0.24173846425118717)(3,0.10895000357641624)(4,0.10319598055131426)(5,0.05661238277164822)
                    (6,0.04439763071464577)(7,0.030057234161169982)(8,0.017700337853604198)(9,0.03435547920150183)(10,0.03274441169901463)
                };
            \addlegendentry{Baum-Welch}
            \addplot[color=red, mark=triangle*] coordinates {
                    (2,0.24173238400408592)(3,0.10970396719908303)(4,0.1035364743891683)(5,0.05161021821642994)
                    (6,0.045517998922194026)(7,0.03775842297782405)(8,0.03170934766211865)(9,0.026481879571591874)(10,0.03236313607742054)
                };
            \addlegendentry{VBEM}
            \nextgroupplot[title={(iii) processed, filtered}]
            \addplot[color=blue, mark=square*] coordinates {
                    (2,0.24171617001181595)(3,0.21864568398172946)(4,0.07278288575658334)(5,0.04060747004339632)
                    (6,0.04479316235402686)(7,0.03965924685588598)(8,0.043354204538529624)(9,0.045809784893832355)(10,0.03770713217178461)
                };
            \addlegendentry{Baum-Welch}
            \addplot[color=red, mark=triangle*] coordinates {
                    (2,0.2417141432627822)(3,0.08531728995804622)(4,0.07082031786880576)(5,0.049742132159070335)
                    (6,0.02606507357238813)(7,0.055526300988869785)(8,0.0423172667614806)(9,0.047731923062633755)(10,0.022112092387832988)
                };
            \addlegendentry{VBEM}
            \nextgroupplot[title={(iv) engineered, filtered}, ylabel={}]
            \addplot[color=blue, mark=square*] coordinates {
                    (2,0.24173846425118717)(3,0.10895003767210443)(4,0.103195980551247)(5,0.056619049825409655)
                    (6,0.04439635841090926)(7,0.030055071535617525)(8,0.017702608835699143)(9,0.03435539569662768)(10,0.03274575988636726)
                };
            \addlegendentry{Baum-Welch}
            \addplot[color=red, mark=triangle*] coordinates {
                    (2,0.24173238400408592)(3,0.109703811625661)(4,0.10353647438899129)(5,0.05160961935059313)
                    (6,0.045519487472239356)(7,0.03775628297810859)(8,0.031709754817909434)(9,0.026482798494198965)(10,0.032363140078487)
                };
            \addlegendentry{VBEM}
        \end{groupplot}
    \end{tikzpicture}
    \caption{Fraud state occupancy for the Baum-Welch and VBEM HMMs across
        $K\in\{2,\ldots,10\}$, batch (top) versus filtered (bottom) posteriors,
        processed (left) versus engineered (right) features. Vertical axis
        scales are independent per panel.}
    \label{fig:baum_welch_vbem_state_occupancy}
\end{figure}

\begin{figure}[H]
    \centering
    \begin{tikzpicture}
        \begin{groupplot}[
                group style={group size=2 by 2, horizontal sep=2cm, vertical sep=2cm,
                        x descriptions at=edge bottom},
                width=0.42\textwidth, height=0.42\textwidth,
                xlabel={Number of hidden states ($K$)}, ylabel={Fraud rate},
                xlabel shift=8pt, ylabel shift=8pt, xtick=data, enlargelimits=0.1,
                ymajorgrids=true, grid style=dashed, scaled y ticks=false,
                xlabel style={font=\small}, ylabel style={font=\small},
                title style={font=\small}, tick label style={font=\footnotesize},
                legend style={font=\scriptsize, inner sep=3pt, row sep=2pt, clip=false, at={(0.98,0.02)}, anchor=south east},
                legend cell align=left, legend image post style={xscale=0.8},
            ]
            \nextgroupplot[title={(i) processed, batch}, xlabel={}]
            \addplot[color=blue, mark=square*] coordinates {
                    (2,0.082)(3,0.090)(4,0.162)(5,0.163)(6,0.216)(7,0.219)(8,0.223)(9,0.222)(10,0.200)
                };
            \addlegendentry{Baum-Welch}
            \addplot[color=red, mark=triangle*] coordinates {
                    (2,0.082)(3,0.135)(4,0.161)(5,0.134)(6,0.152)(7,0.186)(8,0.239)(9,0.217)(10,0.233)
                };
            \addlegendentry{VBEM}
            \nextgroupplot[title={(ii) engineered, batch}, ylabel={}, xlabel={}]
            \addplot[color=blue, mark=square*] coordinates {
                    (2,0.083)(3,0.137)(4,0.134)(5,0.195)(6,0.224)(7,0.211)(8,0.194)(9,0.268)(10,0.267)
                };
            \addlegendentry{Baum-Welch}
            \addplot[color=red, mark=triangle*] coordinates {
                    (2,0.083)(3,0.136)(4,0.134)(5,0.213)(6,0.220)(7,0.253)(8,0.200)(9,0.277)(10,0.275)
                };
            \addlegendentry{VBEM}
            \nextgroupplot[title={(iii) processed, filtered}]
            \addplot[color=blue, mark=square*] coordinates {
                    (2,0.082)(3,0.090)(4,0.162)(5,0.163)(6,0.216)(7,0.219)(8,0.223)(9,0.222)(10,0.200)
                };
            \addlegendentry{Baum-Welch}
            \addplot[color=red, mark=triangle*] coordinates {
                    (2,0.082)(3,0.135)(4,0.161)(5,0.134)(6,0.152)(7,0.186)(8,0.239)(9,0.217)(10,0.233)
                };
            \addlegendentry{VBEM}
            \nextgroupplot[title={(iv) engineered, filtered}, ylabel={}]
            \addplot[color=blue, mark=square*] coordinates {
                    (2,0.083)(3,0.137)(4,0.134)(5,0.195)(6,0.224)(7,0.211)(8,0.194)(9,0.268)(10,0.267)
                };
            \addlegendentry{Baum-Welch}
            \addplot[color=red, mark=triangle*] coordinates {
                    (2,0.083)(3,0.136)(4,0.134)(5,0.213)(6,0.220)(7,0.253)(8,0.200)(9,0.277)(10,0.275)
                };
            \addlegendentry{VBEM}
        \end{groupplot}
    \end{tikzpicture}
    \caption{Fraud state fraud rate ($\eta_{k^*}$) for the Baum-Welch and
        VBEM HMMs across $K\in\{2,\ldots,10\}$, batch (top) versus filtered
        (bottom) posteriors, processed (left) versus engineered (right)
        features.}
    \label{fig:baum_welch_vbem_state_fraud_rate}
\end{figure}
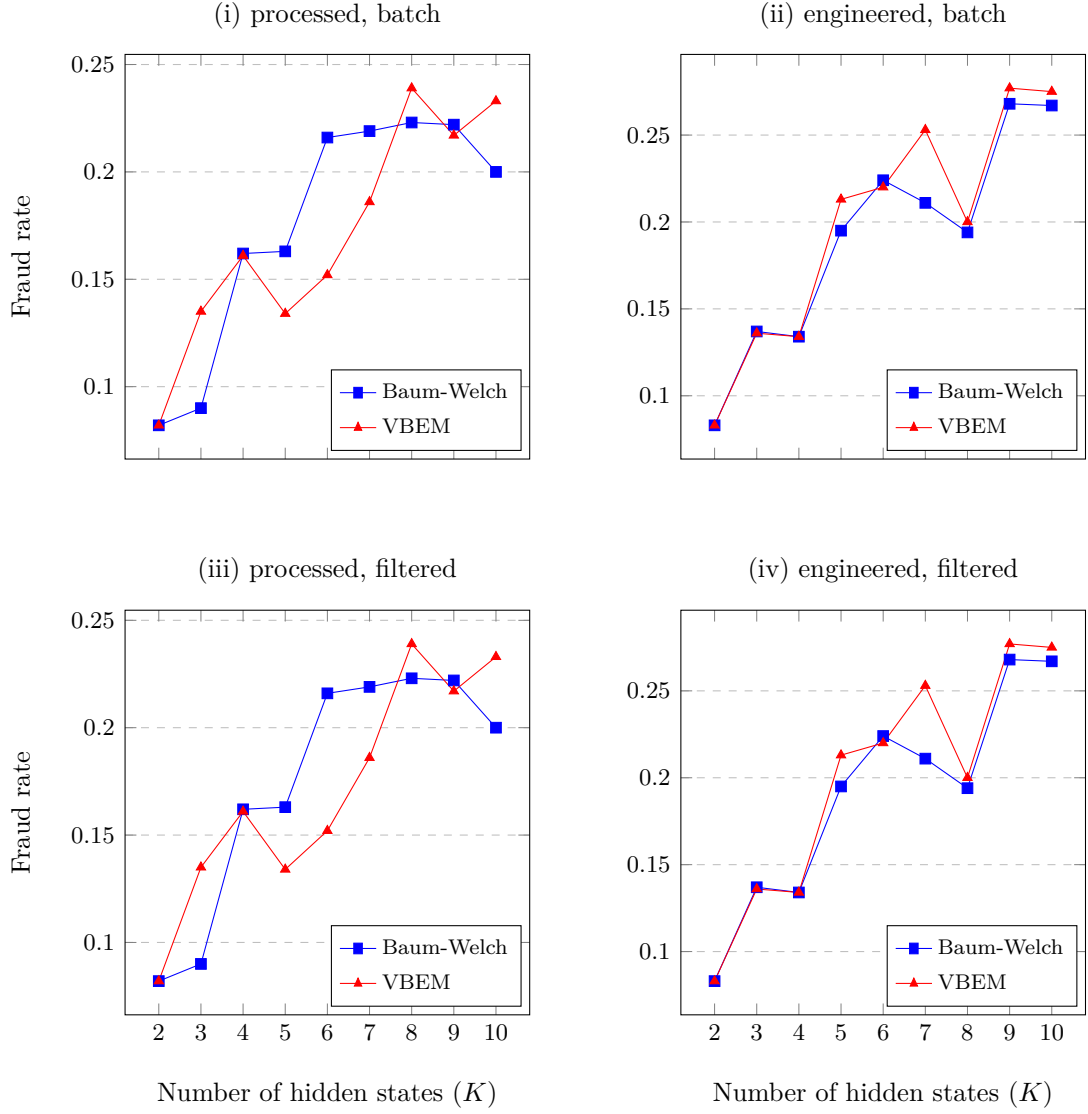

\begin{figure}[H]
    \centering
    \begin{tikzpicture}
        \begin{groupplot}[
                group style={group size=2 by 2, horizontal sep=2cm, vertical sep=2cm,
                        x descriptions at=edge bottom},
                width=0.42\textwidth, height=0.42\textwidth,
                xlabel={Number of hidden states ($K$)}, ylabel={Enrichment},
                xlabel shift=8pt, ylabel shift=8pt, xtick=data, enlargelimits=0.1,
                ymajorgrids=true, grid style=dashed, scaled y ticks=false,
                xlabel style={font=\small}, ylabel style={font=\small},
                title style={font=\small}, tick label style={font=\footnotesize},
                legend style={font=\scriptsize, inner sep=3pt, row sep=2pt, clip=false, at={(0.98,0.02)}, anchor=south east},
                legend cell align=left, legend image post style={xscale=0.8},
            ]
            \nextgroupplot[title={(i) processed, batch}, xlabel={}]
            \addplot[color=blue, mark=square*] coordinates {
                    (2,2.337)(3,2.536)(4,4.596)(5,4.618)(6,6.125)(7,6.194)(8,6.308)(9,6.300)(10,5.651)
                };
            \addlegendentry{Baum-Welch}
            \addplot[color=red, mark=triangle*] coordinates {
                    (2,2.337)(3,3.824)(4,4.549)(5,3.809)(6,4.311)(7,5.261)(8,6.780)(9,6.156)(10,6.599)
                };
            \addlegendentry{VBEM}
            \nextgroupplot[title={(ii) engineered, batch}, ylabel={}, xlabel={}]
            \addplot[color=blue, mark=square*] coordinates {
                    (2,2.343)(3,3.877)(4,3.790)(5,5.513)(6,6.358)(7,5.973)(8,5.484)(9,7.598)(10,7.569)
                };
            \addlegendentry{Baum-Welch}
            \addplot[color=red, mark=triangle*] coordinates {
                    (2,2.344)(3,3.858)(4,3.784)(5,6.043)(6,6.242)(7,7.158)(8,5.652)(9,7.845)(10,7.800)
                };
            \addlegendentry{VBEM}
            \nextgroupplot[title={(iii) processed, filtered}]
            \addplot[color=blue, mark=square*] coordinates {
                    (2,2.337)(3,2.536)(4,4.596)(5,4.618)(6,6.125)(7,6.194)(8,6.308)(9,6.300)(10,5.652)
                };
            \addlegendentry{Baum-Welch}
            \addplot[color=red, mark=triangle*] coordinates {
                    (2,2.337)(3,3.824)(4,4.549)(5,3.809)(6,4.311)(7,5.261)(8,6.779)(9,6.156)(10,6.599)
                };
            \addlegendentry{VBEM}
            \nextgroupplot[title={(iv) engineered, filtered}, ylabel={}]
            \addplot[color=blue, mark=square*] coordinates {
                    (2,2.343)(3,3.877)(4,3.790)(5,5.513)(6,6.358)(7,5.974)(8,5.484)(9,7.598)(10,7.568)
                };
            \addlegendentry{Baum-Welch}
            \addplot[color=red, mark=triangle*] coordinates {
                    (2,2.344)(3,3.858)(4,3.784)(5,6.043)(6,6.242)(7,7.159)(8,5.652)(9,7.845)(10,7.800)
                };
            \addlegendentry{VBEM}
        \end{groupplot}
    \end{tikzpicture}
    \caption{Fraud state enrichment for the Baum-Welch and VBEM HMMs across
        $K\in\{2,\ldots,10\}$, batch (top) versus filtered (bottom) posteriors,
        processed (left) versus engineered (right) features. The near-identical
        top and bottom rows across all three figures are the visual counterpart
        of the batch/filtered agreement reported numerically throughout.}
    \label{fig:baum_welch_vbem_state_enrichment}
\end{figure}
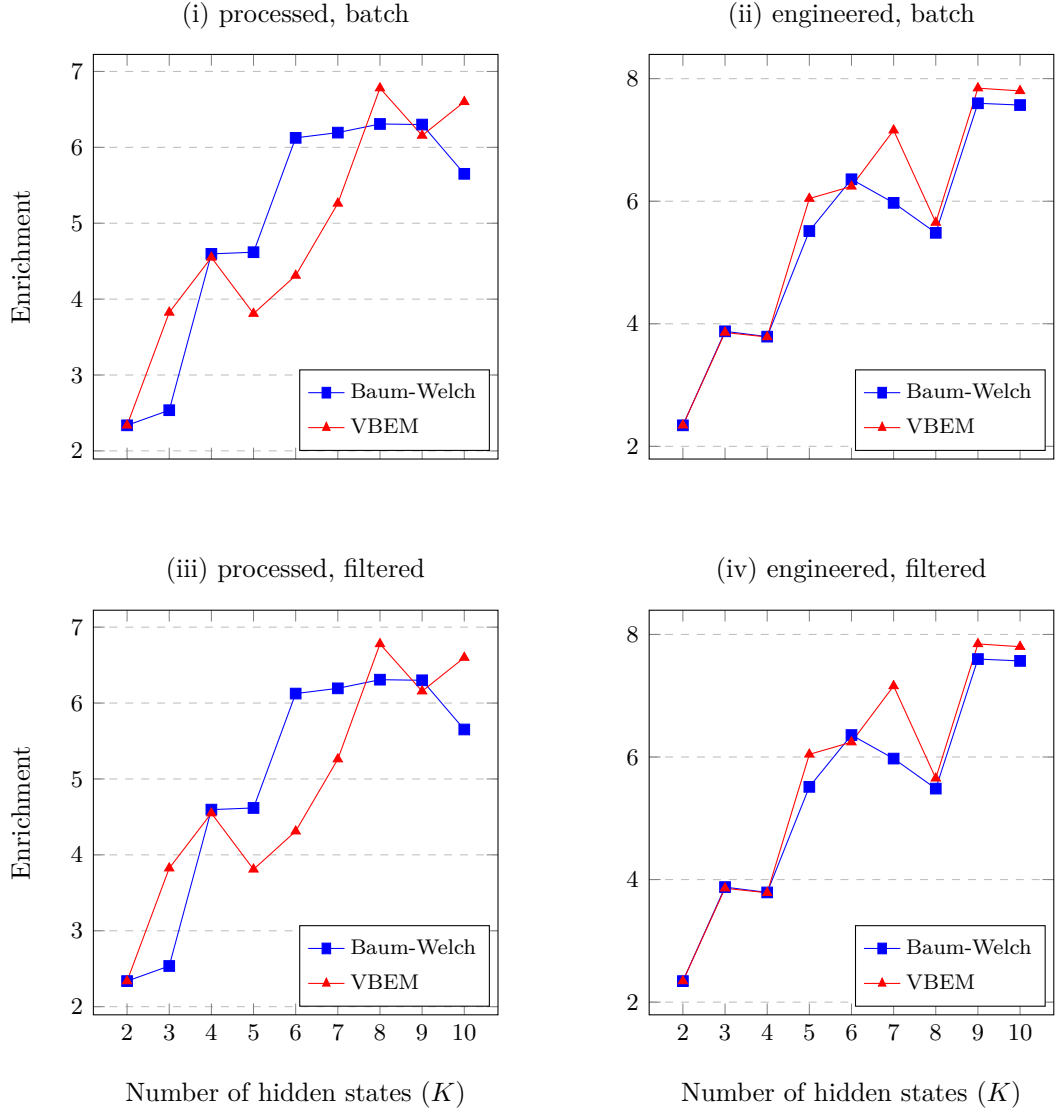

\subsection{State Diagnostics: Neural VBEM HMM}
\label{sec:appc-nvbem-diagnostics}

\begin{table}[H]
    \centering
    \small
    \setstretch{1.3}
    \setlength{\aboverulesep}{0.7ex}
    \setlength{\belowrulesep}{0.7ex}
    \setlength{\tabcolsep}{3.2pt}
    \caption{Neural VBEM HMM ($d_z=32$): fraud-state occupancy, fraud rate,
        and enrichment under batch posteriors, $K\in\{2,\ldots,10\}$, processed (p) and engineered (e) features.}
    \label{tab:appc-nvbem32-batch}
    \begin{tabular}{cS[table-format=1.3]S[table-format=1.3]S[table-format=1.3]S[table-format=1.3]S[table-format=1.3]S[table-format=1.3]}
        \toprule
        {\textbf{K}} & {\textbf{Occup. (p)}} & {\textbf{Occup. (e)}} & {\textbf{F. Rate (p)}} & {\textbf{F. Rate (e)}} & {\textbf{Enrich. (p)}} & {\textbf{Enrich. (e)}} \\
        \midrule
        2            & 0.705                 & 0.441                 & 0.049                  & 0.055                  & 1.386                  & 1.560                  \\
        3            & 0.229                 & 0.145                 & 0.136                  & 0.198                  & 3.860                  & 5.617                  \\
        4            & 0.179                 & 0.128                 & 0.165                  & 0.204                  & 4.675                  & 5.778                  \\
        5            & 0.130                 & 0.119                 & 0.215                  & 0.187                  & 6.079                  & 5.300                  \\
        6            & 0.059                 & 0.122                 & 0.340                  & 0.206                  & 9.644                  & 5.843                  \\
        7            & 0.053                 & 0.075                 & 0.363                  & 0.253                  & 10.292                 & 7.157                  \\
        8            & \textbf{0.044}        & \textbf{0.037}        & \textbf{0.425}         & \textbf{0.461}         & \textbf{12.035}        & \textbf{13.052}        \\
        9            & 0.046                 & 0.042                 & 0.366                  & 0.288                  & 10.365                 & 8.163                  \\
        10           & 0.046                 & 0.067                 & 0.377                  & 0.317                  & 10.676                 & 8.986                  \\
        \bottomrule
    \end{tabular}
\end{table}

\begin{table}[H]
    \centering
    \small
    \setstretch{1.3}
    \setlength{\aboverulesep}{0.7ex}
    \setlength{\belowrulesep}{0.7ex}
    \setlength{\tabcolsep}{3.2pt}
    \caption{Neural VBEM HMM ($d_z=64$): fraud-state occupancy, fraud rate,
        and enrichment under batch posteriors, $K\in\{2,\ldots,10\}$, processed (p) and engineered (e) features.}
    \label{tab:appc-nvbem64-batch}
    \begin{tabular}{cS[table-format=1.3]S[table-format=1.3]S[table-format=1.3]S[table-format=1.3]S[table-format=1.3]S[table-format=1.3]}
        \toprule
        {\textbf{K}} & {\textbf{Occup. (p)}} & {\textbf{Occup. (e)}} & {\textbf{F. Rate (p)}} & {\textbf{F. Rate (e)}} & {\textbf{Enrich. (p)}} & {\textbf{Enrich. (e)}} \\
        \midrule
        2            & 0.391                 & 0.633                 & 0.044                  & 0.054                  & 1.240                  & 1.530                  \\
        3            & 0.384                 & 0.106                 & 0.074                  & 0.243                  & 2.100                  & 6.891                  \\
        4            & 0.120                 & 0.083                 & 0.215                  & 0.307                  & 6.083                  & 8.686                  \\
        5            & 0.116                 & 0.085                 & 0.226                  & 0.285                  & 6.411                  & 8.074                  \\
        6            & 0.065                 & 0.074                 & 0.279                  & 0.314                  & 7.913                  & 8.887                  \\
        7            & 0.063                 & 0.074                 & 0.275                  & 0.239                  & 7.794                  & 6.776                  \\
        8            & 0.066                 & 0.069                 & 0.246                  & 0.228                  & 6.955                  & 6.472                  \\
        9            & 0.040                 & \textbf{0.038}        & 0.384                  & \textbf{0.509}         & 10.883                 & \textbf{14.427}        \\
        10           & \textbf{0.034}        & 0.043                 & \textbf{0.481}         & 0.462                  & \textbf{13.612}        & 13.087                 \\
        \bottomrule
    \end{tabular}
\end{table}

\begin{table}[H]
    \centering
    \small
    \setstretch{1.3}
    \setlength{\aboverulesep}{0.7ex}
    \setlength{\belowrulesep}{0.7ex}
    \setlength{\tabcolsep}{3.2pt}
    \caption{Neural VBEM HMM ($d_z=128$): fraud-state occupancy, fraud rate,
        and enrichment under batch posteriors, $K\in\{2,\ldots,10\}$, processed (p) and engineered (e) features.}
    \label{tab:appc-nvbem128-batch}
    \begin{tabular}{cS[table-format=1.3]S[table-format=1.3]S[table-format=1.3]S[table-format=1.3]S[table-format=1.3]S[table-format=1.3]}
        \toprule
        {\textbf{K}} & {\textbf{Occup. (p)}} & {\textbf{Occup. (e)}} & {\textbf{F. Rate (p)}} & {\textbf{F. Rate (e)}} & {\textbf{Enrich. (p)}} & {\textbf{Enrich. (e)}} \\
        \midrule
        2            & 0.360                 & 0.448                 & 0.053                  & 0.059                  & 1.493                  & 1.677                  \\
        3            & 0.254                 & 0.181                 & 0.051                  & 0.089                  & 1.442                  & 2.521                  \\
        4            & 0.111                 & 0.113                 & 0.237                  & 0.244                  & 6.700                  & 6.903                  \\
        5            & 0.071                 & 0.125                 & 0.264                  & 0.208                  & 7.489                  & 5.888                  \\
        6            & 0.066                 & 0.082                 & 0.247                  & 0.313                  & 6.986                  & 8.864                  \\
        7            & 0.072                 & 0.088                 & 0.206                  & 0.274                  & 5.847                  & 7.774                  \\
        8            & 0.062                 & 0.110                 & 0.239                  & 0.202                  & 6.761                  & 5.732                  \\
        9            & 0.056                 & \textbf{0.038}        & 0.246                  & \textbf{0.551}         & 6.977                  & \textbf{15.594}        \\
        10           & \textbf{0.046}        & 0.076                 & \textbf{0.303}         & 0.244                  & \textbf{8.594}         & 6.911                  \\
        \bottomrule
    \end{tabular}
\end{table}

The $d_z=32$ and $d_z=128$ configurations both show markedly more erratic
enrichment trajectories than $d_z=64$: $d_z=32$ engineered peaks early, at
$K=8$ ($13.05\times$), then declines at $K=9$--$10$, as $d_z=128$
engineered stays modest through $K=8$ before a single-step spike to
$15.59\times$ at $K=9$, the highest enrichment recorded anywhere,
immediately followed by a collapse to $6.91\times$ at $K=10$,
a swing of $8.7\times$ in one model-order step. $d_z=64$'s corresponding
transition ($14.43\times\to13.09\times$) is, by comparison, the most
stable of the three, which is the main reason it was carried forward as
the representative configuration (\cref{sec:model-selection}) over the
$d_z=128$ configuration's higher single-point peak, a result that
depends on hitting one exact $K$ and reverses sharply either side of it, as
a weaker basis for a production model than one that degrades
gracefully.

Visual comparison across the three encoder widths follows for all three
diagnostics, one row per $d_z$ configuration, processed features on the
left and engineered on the right.

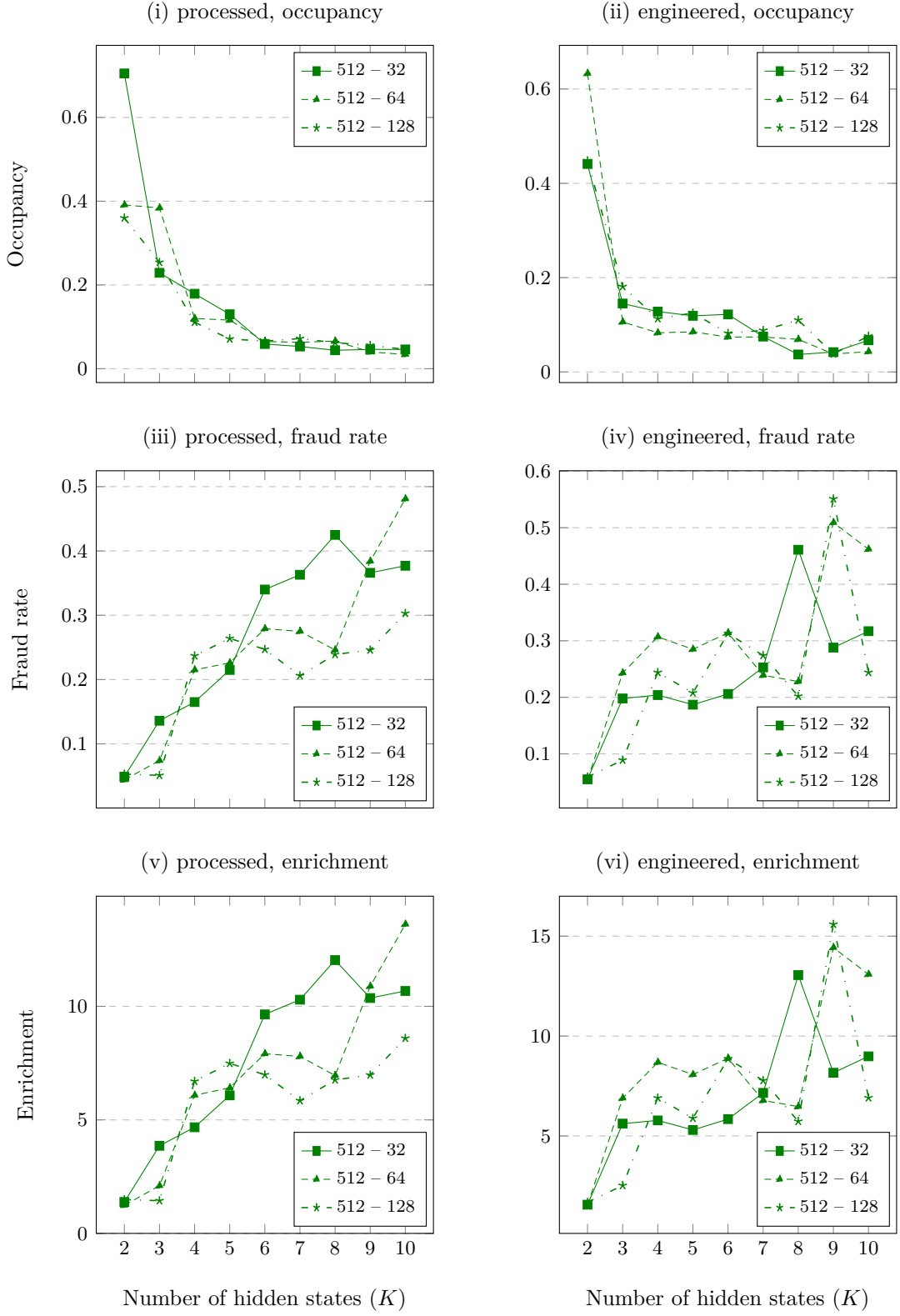
\begin{figure}[H]
    \centering
    \begin{tikzpicture}
        \begin{groupplot}[
                group style={group size=2 by 3, horizontal sep=2cm, vertical sep=1.4cm,
                        x descriptions at=edge bottom},
                width=0.42\textwidth, height=0.42\textwidth,
                xlabel={Number of hidden states ($K$)},
                xlabel shift=8pt, ylabel shift=8pt, xtick=data, enlargelimits=0.1,
                ymajorgrids=true, grid style=dashed, scaled y ticks=false,
                xlabel style={font=\small}, ylabel style={font=\small},
                title style={font=\small}, tick label style={font=\footnotesize},
                legend style={font=\scriptsize, inner sep=3pt, row sep=2pt, clip=false, at={(0.98,0.98)}, anchor=north east},
                legend cell align=left, legend image post style={xscale=0.8},
            ]

            \nextgroupplot[title={(i) processed, occupancy}, ylabel={Occupancy}, xlabel={}]
            \addplot[color=green!50!black, mark=square*] coordinates {
                    (2,0.705)(3,0.229)(4,0.179)(5,0.130)(6,0.059)(7,0.053)(8,0.044)(9,0.046)(10,0.046)
                };
            \addlegendentry{512 -- 32}
            \addplot[color=green!50!black, mark=triangle*, densely dashed] coordinates {
                    (2,0.391)(3,0.384)(4,0.120)(5,0.116)(6,0.065)(7,0.063)(8,0.066)(9,0.040)(10,0.034)
                };
            \addlegendentry{512 -- 64}
            \addplot[color=green!50!black, mark=star, loosely dashdotted, semithick] coordinates {
                    (2,0.360)(3,0.254)(4,0.111)(5,0.071)(6,0.066)(7,0.072)(8,0.062)(9,0.056)(10,0.046)
                };
            \addlegendentry{512 -- 128}

            \nextgroupplot[title={(ii) engineered, occupancy}, ylabel={}, xlabel={}]
            \addplot[color=green!50!black, mark=square*] coordinates {
                    (2,0.441)(3,0.145)(4,0.128)(5,0.119)(6,0.122)(7,0.075)(8,0.037)(9,0.042)(10,0.067)
                };
            \addlegendentry{512 -- 32}
            \addplot[color=green!50!black, mark=triangle*, densely dashed] coordinates {
                    (2,0.633)(3,0.106)(4,0.083)(5,0.085)(6,0.074)(7,0.074)(8,0.069)(9,0.038)(10,0.043)
                };
            \addlegendentry{512 -- 64}
            \addplot[color=green!50!black, mark=star, loosely dashdotted, semithick] coordinates {
                    (2,0.448)(3,0.181)(4,0.113)(5,0.125)(6,0.082)(7,0.088)(8,0.110)(9,0.038)(10,0.076)
                };
            \addlegendentry{512 -- 128}

            \nextgroupplot[title={(iii) processed, fraud rate}, ylabel={Fraud rate}, xlabel={}, legend style={at={(0.98,0.02)}, anchor=south east}]
            \addplot[color=green!50!black, mark=square*] coordinates {
                    (2,0.049)(3,0.136)(4,0.165)(5,0.215)(6,0.340)(7,0.363)(8,0.425)(9,0.366)(10,0.377)
                };
            \addlegendentry{512 -- 32}
            \addplot[color=green!50!black, mark=triangle*, densely dashed] coordinates {
                    (2,0.044)(3,0.074)(4,0.215)(5,0.226)(6,0.279)(7,0.275)(8,0.246)(9,0.384)(10,0.481)
                };
            \addlegendentry{512 -- 64}
            \addplot[color=green!50!black, mark=star, loosely dashdotted, semithick] coordinates {
                    (2,0.053)(3,0.051)(4,0.237)(5,0.264)(6,0.247)(7,0.206)(8,0.239)(9,0.246)(10,0.303)
                };
            \addlegendentry{512 -- 128}

            \nextgroupplot[title={(iv) engineered, fraud rate}, ylabel={}, xlabel={}, legend style={at={(0.98,0.02)}, anchor=south east}]
            \addplot[color=green!50!black, mark=square*] coordinates {
                    (2,0.055)(3,0.198)(4,0.204)(5,0.187)(6,0.206)(7,0.253)(8,0.461)(9,0.288)(10,0.317)
                };
            \addlegendentry{512 -- 32}
            \addplot[color=green!50!black, mark=triangle*, densely dashed] coordinates {
                    (2,0.054)(3,0.243)(4,0.307)(5,0.285)(6,0.314)(7,0.239)(8,0.228)(9,0.509)(10,0.462)
                };
            \addlegendentry{512 -- 64}
            \addplot[color=green!50!black, mark=star, loosely dashdotted, semithick] coordinates {
                    (2,0.059)(3,0.089)(4,0.244)(5,0.208)(6,0.313)(7,0.274)(8,0.202)(9,0.551)(10,0.244)
                };
            \addlegendentry{512 -- 128}

            \nextgroupplot[title={(v) processed, enrichment}, ylabel={Enrichment}, legend style={at={(0.98,0.02)}, anchor=south east}]
            \addplot[color=green!50!black, mark=square*] coordinates {
                    (2,1.386)(3,3.860)(4,4.675)(5,6.079)(6,9.644)(7,10.292)(8,12.035)(9,10.365)(10,10.676)
                };
            \addlegendentry{512 -- 32}
            \addplot[color=green!50!black, mark=triangle*, densely dashed] coordinates {
                    (2,1.240)(3,2.100)(4,6.083)(5,6.411)(6,7.913)(7,7.794)(8,6.955)(9,10.883)(10,13.612)
                };
            \addlegendentry{512 -- 64}
            \addplot[color=green!50!black, mark=star, loosely dashdotted, semithick] coordinates {
                    (2,1.493)(3,1.442)(4,6.700)(5,7.489)(6,6.986)(7,5.847)(8,6.761)(9,6.977)(10,8.594)
                };
            \addlegendentry{512 -- 128}

            \nextgroupplot[title={(vi) engineered, enrichment}, ylabel={}, legend style={at={(0.98,0.02)}, anchor=south east}]
            \addplot[color=green!50!black, mark=square*] coordinates {
                    (2,1.560)(3,5.617)(4,5.778)(5,5.300)(6,5.843)(7,7.157)(8,13.052)(9,8.163)(10,8.986)
                };
            \addlegendentry{512 -- 32}
            \addplot[color=green!50!black, mark=triangle*, densely dashed] coordinates {
                    (2,1.530)(3,6.891)(4,8.686)(5,8.074)(6,8.887)(7,6.776)(8,6.472)(9,14.427)(10,13.087)
                };
            \addlegendentry{512 -- 64}
            \addplot[color=green!50!black, mark=star, loosely dashdotted, semithick] coordinates {
                    (2,1.677)(3,2.521)(4,6.903)(5,5.888)(6,8.864)(7,7.774)(8,5.732)(9,15.594)(10,6.911)
                };
            \addlegendentry{512 -- 128}

        \end{groupplot}
    \end{tikzpicture}
    \caption{Fraud state occupancy, fraud rate, and enrichment for the
        Neural VBEM HMM ($d_h=512$) across $K\in\{2,\ldots,10\}$ and latent
        dimension $d_z\in\{32,64,128\}$, processed (left) versus engineered
        (right) features, batch posteriors.}
    \label{fig:neural_vbem_state_diagnostics}
\end{figure}

\subsection{Transition Structure and State Persistence}
\label{app:stickiness}

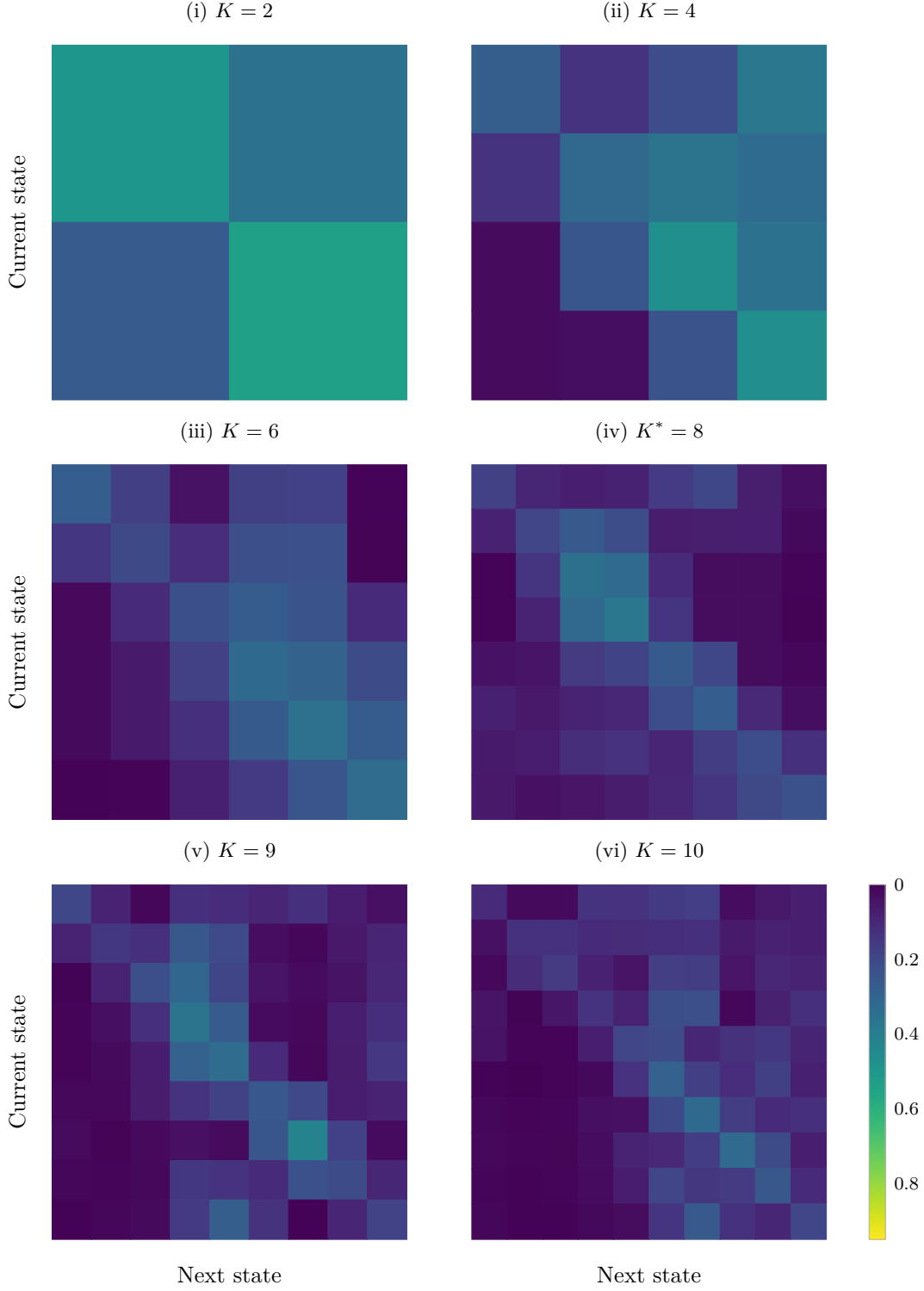
\begin{figure}[H]
    \centering
    \begin{tikzpicture}
        \begin{groupplot}[
                group style={group size=2 by 3, horizontal sep=1cm, vertical sep=1cm,
                        x descriptions at=edge bottom, y descriptions at=edge left},
                width=0.33\textwidth, height=0.33\textwidth, enlargelimits=false,
                axis line style={draw=black!30}, scale only axis,
                xlabel={Next state}, ylabel={Current state},
                xlabel shift=8pt, ylabel shift=8pt,
                xlabel style={font=\small}, ylabel style={font=\small},
                xtick=\empty, ytick=\empty, tick style={draw=none},
                title style={font=\footnotesize}, y dir=reverse,
                colormap/viridis, point meta min=0, point meta max=0.95,
            ]
            \nextgroupplot[title={(i) $K=2$}]
            \addplot[matrix plot*, mesh/cols=2, shader=flat, point meta=explicit] coordinates{
                    (1,1)[0.748] (2,1)[0.252] (1,2)[0.079] (2,2)[0.921]
                };
            \nextgroupplot[title={(ii) $K=4$}]
            \addplot[matrix plot*, mesh/cols=4, shader=flat, point meta=explicit] coordinates{
                    (1,1)[0.386](2,1)[0.210](3,1)[0.028](4,1)[0.376]
                    (1,2)[0.225](2,2)[0.304](3,2)[0.010](4,2)[0.461]
                    (1,3)[0.015](2,3)[0.009](3,3)[0.952](4,3)[0.024]
                    (1,4)[0.025](2,4)[0.050](3,4)[0.003](4,4)[0.921]
                };
            \nextgroupplot[title={(iii) $K=6$}]
            \addplot[matrix plot*, mesh/cols=6, shader=flat, point meta=explicit] coordinates{
                    (1,1)[0.355](2,1)[0.231](3,1)[0.077](4,1)[0.006](5,1)[0.323](6,1)[0.008]
                    (1,2)[0.219](2,2)[0.306](3,2)[0.084](4,2)[0.004](5,2)[0.379](6,2)[0.007]
                    (1,3)[0.022](2,3)[0.046](3,3)[0.394](4,3)[0.001](5,3)[0.533](6,3)[0.003]
                    (1,4)[0.006](2,4)[0.010](3,4)[0.006](4,4)[0.530](5,4)[0.021](6,4)[0.427]
                    (1,5)[0.024](2,5)[0.054](3,5)[0.187](4,5)[0.001](5,5)[0.732](6,5)[0.002]
                    (1,6)[0.004](2,6)[0.010](3,6)[0.007](4,6)[0.306](5,6)[0.019](6,6)[0.655]
                };
            \nextgroupplot[title={(iv) $K^*=8$}]
            \addplot[matrix plot*, mesh/cols=8, shader=flat, point meta=explicit] coordinates{
                    (1,1)[0.393](2,1)[0.002](3,1)[0.004](4,1)[0.008](5,1)[0.004](6,1)[0.547](7,1)[0.007](8,1)[0.036]
                    (1,2)[0.002](2,2)[0.327](3,2)[0.064](4,2)[0.241](5,2)[0.094](6,2)[0.010](7,2)[0.244](8,2)[0.018]
                    (1,3)[0.001](2,3)[0.014](3,3)[0.382](4,3)[0.364](5,3)[0.184](6,3)[0.002](7,3)[0.049](8,3)[0.004]
                    (1,4)[0.000](2,4)[0.017](3,4)[0.157](4,4)[0.491](5,4)[0.264](6,4)[0.003](7,4)[0.064](8,4)[0.004]
                    (1,5)[0.000](2,5)[0.014](3,5)[0.178](4,5)[0.452](5,5)[0.300](6,5)[0.002](7,5)[0.050](8,5)[0.003]
                    (1,6)[0.157](2,6)[0.004](3,6)[0.003](4,6)[0.015](5,6)[0.006](6,6)[0.750](7,6)[0.010](8,6)[0.054]
                    (1,7)[0.002](2,7)[0.166](3,7)[0.070](4,7)[0.277](5,7)[0.112](6,7)[0.008](7,7)[0.344](8,7)[0.023]
                    (1,8)[0.058](2,8)[0.031](3,8)[0.032](4,8)[0.109](5,8)[0.042](6,8)[0.220](7,8)[0.098](8,8)[0.411]
                };
            \nextgroupplot[title={(v) $K=9$}]
            \addplot[matrix plot*, mesh/cols=9, shader=flat, point meta=explicit] coordinates{
                    (1,1)[0.464](2,1)[0.005](3,1)[0.003](4,1)[0.008](5,1)[0.014](6,1)[0.007](7,1)[0.366](8,1)[0.130](9,1)[0.004]
                    (1,2)[0.001](2,2)[0.341](3,2)[0.011](4,2)[0.034](5,2)[0.448](6,2)[0.004](7,2)[0.012](8,2)[0.003](9,2)[0.146]
                    (1,3)[0.001](2,3)[0.012](3,3)[0.241](4,3)[0.218](5,3)[0.330](6,3)[0.062](7,3)[0.044](8,3)[0.002](9,3)[0.089]
                    (1,4)[0.001](2,4)[0.010](3,4)[0.098](4,4)[0.352](5,4)[0.357](6,4)[0.037](7,4)[0.054](8,4)[0.003](9,4)[0.089]
                    (1,5)[0.000](2,5)[0.020](3,5)[0.013](4,5)[0.055](5,5)[0.675](6,5)[0.005](7,5)[0.010](8,5)[0.001](9,5)[0.222]
                    (1,6)[0.003](2,6)[0.009](3,6)[0.044](4,6)[0.175](5,6)[0.257](6,6)[0.397](7,6)[0.047](8,6)[0.003](9,6)[0.064]
                    (1,7)[0.051](2,7)[0.004](3,7)[0.018](4,7)[0.056](5,7)[0.065](6,7)[0.013](7,7)[0.563](8,7)[0.212](9,7)[0.018]
                    (1,8)[0.040](2,8)[0.002](3,8)[0.002](4,8)[0.009](5,8)[0.018](6,8)[0.002](7,8)[0.437](8,8)[0.486](9,8)[0.004]
                    (1,9)[0.000](2,9)[0.022](3,9)[0.012](4,9)[0.046](5,9)[0.533](6,9)[0.004](7,9)[0.008](8,9)[0.001](9,9)[0.374]
                };
            \nextgroupplot[title={(vi) $K=10$}, colorbar,
                colorbar style={at={(1.12,0.5)}, anchor=west, ytick={0,0.2,0.4,0.6,0.8},
                        yticklabel style={font=\scriptsize}, width=0.25cm, height=0.33\textwidth}]
            \addplot[matrix plot*, mesh/cols=10, shader=flat, point meta=explicit] coordinates{
                    (1,1)[0.331](2,1)[0.008](3,1)[0.011](4,1)[0.010](5,1)[0.207](6,1)[0.002](7,1)[0.305](8,1)[0.010](9,1)[0.036](10,1)[0.080]
                    (1,2)[0.086](2,2)[0.015](3,2)[0.055](4,2)[0.012](5,2)[0.326](6,2)[0.001](7,2)[0.334](8,2)[0.035](9,2)[0.032](10,2)[0.105]
                    (1,3)[0.043](2,3)[0.005](3,3)[0.456](4,3)[0.007](5,3)[0.100](6,3)[0.040](7,3)[0.122](8,3)[0.019](9,3)[0.177](10,3)[0.032]
                    (1,4)[0.017](2,4)[0.001](3,4)[0.002](4,4)[0.189](5,4)[0.050](6,4)[0.001](7,4)[0.535](8,4)[0.002](9,4)[0.009](10,4)[0.195]
                    (1,5)[0.165](2,5)[0.011](3,5)[0.010](4,5)[0.010](5,5)[0.303](6,5)[0.002](7,5)[0.364](8,5)[0.005](9,5)[0.041](10,5)[0.090]
                    (1,6)[0.004](2,6)[0.000](3,6)[0.010](4,6)[0.000](5,6)[0.007](6,6)[0.479](7,6)[0.017](8,6)[0.008](9,6)[0.470](10,6)[0.004]
                    (1,7)[0.015](2,7)[0.002](3,7)[0.001](4,7)[0.016](5,7)[0.055](6,7)[0.001](7,7)[0.674](8,7)[0.001](9,7)[0.007](10,7)[0.228]
                    (1,8)[0.041](2,8)[0.003](3,8)[0.021](4,8)[0.005](5,8)[0.068](6,8)[0.033](7,8)[0.127](8,8)[0.490](9,8)[0.172](10,8)[0.040]
                    (1,9)[0.021](2,9)[0.002](3,9)[0.018](4,9)[0.001](5,9)[0.041](6,9)[0.210](7,9)[0.049](8,9)[0.015](9,9)[0.629](10,9)[0.013]
                    (1,10)[0.014](2,10)[0.002](3,10)[0.001](4,10)[0.018](5,10)[0.045](6,10)[0.001](7,10)[0.527](8,10)[0.001](9,10)[0.006](10,10)[0.385]
                };
        \end{groupplot}
    \end{tikzpicture}
    \caption{Row-normalised transition matrices $A\in[0,1]^{K\times K}$ for
        the Baum-Welch HMM, engineered features, $K\in\{2,4,6,8,9,10\}$. Panel
        (iv) is the selected order $K^*=8$; cell $(i,j)$ is $A_{ij}=
            \mathbb{P}(q_{t+1}=j\mid q_t=i)$, rows sum to 1.}
    \label{fig:appc-bw-transitions}
\end{figure}

At $K^*=8$, the fraud-associated state (state 8) has self-transition
$A_{8,8}=0.411$, denoting intermediate stickiness, more persistent than a purely
episodic regime but well short of the dominant legitimate states (e.g.\
$A_{6,6}=0.750$), meaning that once the model places an account in the
fraud regime, there is a better-than-even chance it remains there at the
next transaction, giving the fraud signal temporal persistence. At $K=10$, this structure
reorganises, as the dominant legitimate state's persistence
($A_{6,6}=0.750$ at $K^*=8$) splits across two new attractors
($A_{7,7}=0.674$, $A_{9,9}=0.629$), consistent with the BIC increase and
the agreement-criterion failure both observed at $K=10$.

\begin{figure}[H]
    \centering
    \begin{tikzpicture}
        \begin{groupplot}[
                group style={group size=2 by 3, horizontal sep=1cm, vertical sep=1cm,
                        x descriptions at=edge bottom, y descriptions at=edge left},
                width=0.33\textwidth, height=0.33\textwidth, enlargelimits=false,
                axis line style={draw=black!30}, scale only axis,
                xlabel={Next state}, ylabel={Current state},
                xlabel shift=8pt, ylabel shift=8pt,
                xlabel style={font=\small}, ylabel style={font=\small},
                xtick=\empty, ytick=\empty, tick style={draw=none},
                title style={font=\footnotesize}, y dir=reverse,
                colormap/viridis, point meta min=0, point meta max=0.95,
            ]
            \nextgroupplot[title={(i) $K=2$}]
            \addplot[matrix plot*, mesh/cols=2, shader=flat, point meta=explicit] coordinates{
                    (1,1)[0.748](2,1)[0.252](1,2)[0.079](2,2)[0.921]
                };
            \nextgroupplot[title={(ii) $K=4$}]
            \addplot[matrix plot*, mesh/cols=4, shader=flat, point meta=explicit] coordinates{
                    (1,1)[0.953](2,1)[0.010](3,1)[0.016](4,1)[0.022]
                    (1,2)[0.011](2,2)[0.306](3,2)[0.230](4,2)[0.454]
                    (1,3)[0.028](2,3)[0.209](3,3)[0.388](4,3)[0.374]
                    (1,4)[0.003](2,4)[0.050](3,4)[0.026](4,4)[0.921]
                };
            \nextgroupplot[title={(iii) $K=6$}]
            \addplot[matrix plot*, mesh/cols=6, shader=flat, point meta=explicit] coordinates{
                    (1,1)[0.525](2,1)[0.417](3,1)[0.034](4,1)[0.002](5,1)[0.013](6,1)[0.010]
                    (1,2)[0.306](2,2)[0.655](3,2)[0.023](4,2)[0.002](5,2)[0.008](6,2)[0.005]
                    (1,3)[0.002](2,3)[0.002](3,3)[0.921](4,3)[0.000](5,3)[0.043](6,3)[0.032]
                    (1,4)[0.035](2,4)[0.089](3,4)[0.036](4,4)[0.800](5,4)[0.025](6,4)[0.015]
                    (1,5)[0.006](2,5)[0.008](3,5)[0.458](4,5)[0.001](5,5)[0.267](6,5)[0.260]
                    (1,6)[0.008](2,6)[0.007](3,6)[0.416](4,6)[0.001](5,6)[0.201](6,6)[0.366]
                };
            \nextgroupplot[title={(iv) $K^*=7$}]
            \addplot[matrix plot*, mesh/cols=7, shader=flat, point meta=explicit] coordinates{
                    (1,1)[0.519](2,1)[0.436](3,1)[0.026](4,1)[0.002](5,1)[0.010](6,1)[0.006](7,1)[0.002]
                    (1,2)[0.218](2,2)[0.656](3,2)[0.060](4,2)[0.007](5,2)[0.035](6,2)[0.021](7,2)[0.002]
                    (1,3)[0.001](2,3)[0.007](3,3)[0.921](4,3)[0.011](5,3)[0.043](6,3)[0.016](7,3)[0.000]
                    (1,4)[0.004](2,4)[0.034](3,4)[0.466](4,4)[0.128](5,4)[0.204](6,4)[0.148](7,4)[0.016]
                    (1,5)[0.004](2,5)[0.042](3,5)[0.457](4,5)[0.053](5,5)[0.266](6,5)[0.178](7,5)[0.001]
                    (1,6)[0.005](2,6)[0.035](3,6)[0.393](4,6)[0.043](5,6)[0.177](6,6)[0.347](7,6)[0.001]
                    (1,7)[0.044](2,7)[0.154](3,7)[0.053](4,7)[0.240](5,7)[0.028](6,7)[0.009](7,7)[0.471]
                };
            \nextgroupplot[title={(v) $K=9$}]
            \addplot[matrix plot*, mesh/cols=9, shader=flat, point meta=explicit] coordinates{
                    (1,1)[0.417](2,1)[0.068](3,1)[0.021](4,1)[0.011](5,1)[0.001](6,1)[0.288](7,1)[0.001](8,1)[0.082](9,1)[0.113]
                    (1,2)[0.014](2,2)[0.299](3,2)[0.120](4,2)[0.050](5,2)[0.002](6,2)[0.053](7,2)[0.000](8,2)[0.002](9,2)[0.460]
                    (1,3)[0.022](2,3)[0.200](3,3)[0.340](4,3)[0.027](5,3)[0.002](6,3)[0.045](7,3)[0.001](8,3)[0.004](9,3)[0.359]
                    (1,4)[0.024](2,4)[0.214](3,4)[0.089](4,4)[0.148](5,4)[0.001](6,4)[0.045](7,4)[0.001](8,4)[0.002](9,4)[0.477]
                    (1,5)[0.004](2,5)[0.042](3,5)[0.005](4,5)[0.011](5,5)[0.253](6,5)[0.016](7,5)[0.077](8,5)[0.002](9,5)[0.589]
                    (1,6)[0.074](2,6)[0.051](3,6)[0.014](4,6)[0.010](5,6)[0.001](6,6)[0.606](7,6)[0.002](8,6)[0.164](9,6)[0.077]
                    (1,7)[0.012](2,7)[0.031](3,7)[0.004](4,7)[0.006](5,7)[0.239](6,7)[0.138](7,7)[0.533](8,7)[0.024](9,7)[0.013]
                    (1,8)[0.069](2,8)[0.007](3,8)[0.003](4,8)[0.002](5,8)[0.000](6,8)[0.461](7,8)[0.002](8,8)[0.433](9,8)[0.023]
                    (1,9)[0.002](2,9)[0.051](3,9)[0.007](4,9)[0.009](5,9)[0.003](6,9)[0.009](7,9)[0.000](8,9)[0.001](9,9)[0.918]
                };
            \nextgroupplot[title={(vi) $K=10$}, colorbar,
                colorbar style={at={(1.12,0.5)}, anchor=west, ytick={0,0.2,0.4,0.6,0.8},
                        yticklabel style={font=\scriptsize}, width=0.25cm, height=0.33\textwidth}]
            \addplot[matrix plot*, mesh/cols=10, shader=flat, point meta=explicit] coordinates{
                    (1,1)[0.034](2,1)[0.025](3,1)[0.036](4,1)[0.023](5,1)[0.051](6,1)[0.081](7,1)[0.108](8,1)[0.302](9,1)[0.251](10,1)[0.089]
                    (1,2)[0.001](2,2)[0.463](3,2)[0.005](4,2)[0.001](5,2)[0.004](6,2)[0.006](7,2)[0.458](8,2)[0.013](9,2)[0.027](10,2)[0.023]
                    (1,3)[0.005](2,3)[0.007](3,3)[0.335](4,3)[0.002](5,3)[0.048](6,3)[0.067](7,3)[0.011](8,3)[0.213](9,3)[0.298](10,3)[0.014]
                    (1,4)[0.003](2,4)[0.005](3,4)[0.007](4,4)[0.443](5,4)[0.011](6,4)[0.126](7,4)[0.014](8,4)[0.036](9,4)[0.353](10,4)[0.003]
                    (1,5)[0.005](2,5)[0.005](3,5)[0.112](4,5)[0.002](5,5)[0.140](6,5)[0.088](7,5)[0.009](8,5)[0.266](9,5)[0.363](10,5)[0.010]
                    (1,6)[0.001](2,6)[0.001](3,6)[0.008](4,6)[0.003](5,6)[0.010](6,6)[0.375](7,6)[0.002](8,6)[0.047](9,6)[0.552](10,6)[0.002]
                    (1,7)[0.002](2,7)[0.224](3,7)[0.002](4,7)[0.001](5,7)[0.002](6,7)[0.004](7,7)[0.702](8,7)[0.009](9,7)[0.018](10,7)[0.036]
                    (1,8)[0.005](2,8)[0.004](3,8)[0.047](4,8)[0.002](5,8)[0.061](6,8)[0.079](7,8)[0.008](8,8)[0.416](9,8)[0.367](10,8)[0.012]
                    (1,9)[0.001](2,9)[0.001](3,9)[0.009](4,9)[0.003](5,9)[0.011](6,9)[0.212](7,9)[0.002](8,9)[0.057](9,9)[0.703](10,9)[0.002]
                    (1,10)[0.004](2,10)[0.071](3,10)[0.021](4,10)[0.002](5,10)[0.022](6,10)[0.029](7,10)[0.234](8,10)[0.097](9,10)[0.106](10,10)[0.415]
                };
        \end{groupplot}
    \end{tikzpicture}
    \caption{Row-normalised transition matrices for the VBEM HMM, engineered
        features, $K\in\{2,4,6,7,9,10\}$. Panel (iv) is the selected order
        $K^*=7$; colour scale identical to \cref{fig:appc-bw-transitions}.}
    \label{fig:appc-vbem-transitions}
\end{figure}

At $K=2$, VBEM's transition matrix is virtually identical to
Baum-Welch's ($A_{1,1}=0.748$, $A_{2,2}=0.921$), consistent with the
state diagnostics, as abundant evidence per state leaves the NIW prior no
room to move the posterior from the maximum-likelihood estimate. By
$K=4$, the two tiers diverge structurally, where VBEM's state 1 reaches
$A_{1,1}=0.953$, the single highest self-transition recorded anywhere in
this work, against Baum-Welch's $0.952$ at a different state and model
order, and by $K^*=7$, a single dominant legitimate state
($A_{3,3}=0.921$) organises the whole matrix, with the fraud state at
$A_{1,1}=0.519$, essentially the same intermediate stickiness as
Baum-Welch's fraud state despite the two tiers using entirely different
inference procedures. At $K=10$, that single dominant anchor splits into
two ($A_{7,7}=0.702$, $A_{9,9}=0.703$), the same bifurcation pattern seen
in Baum-Welch, coinciding with the sharp ELBO increase noted in
\cref{fig:appc-vbem-elbo}: higher marginal likelihood bought by
splitting an existing regime as opposed to discovering a new one.

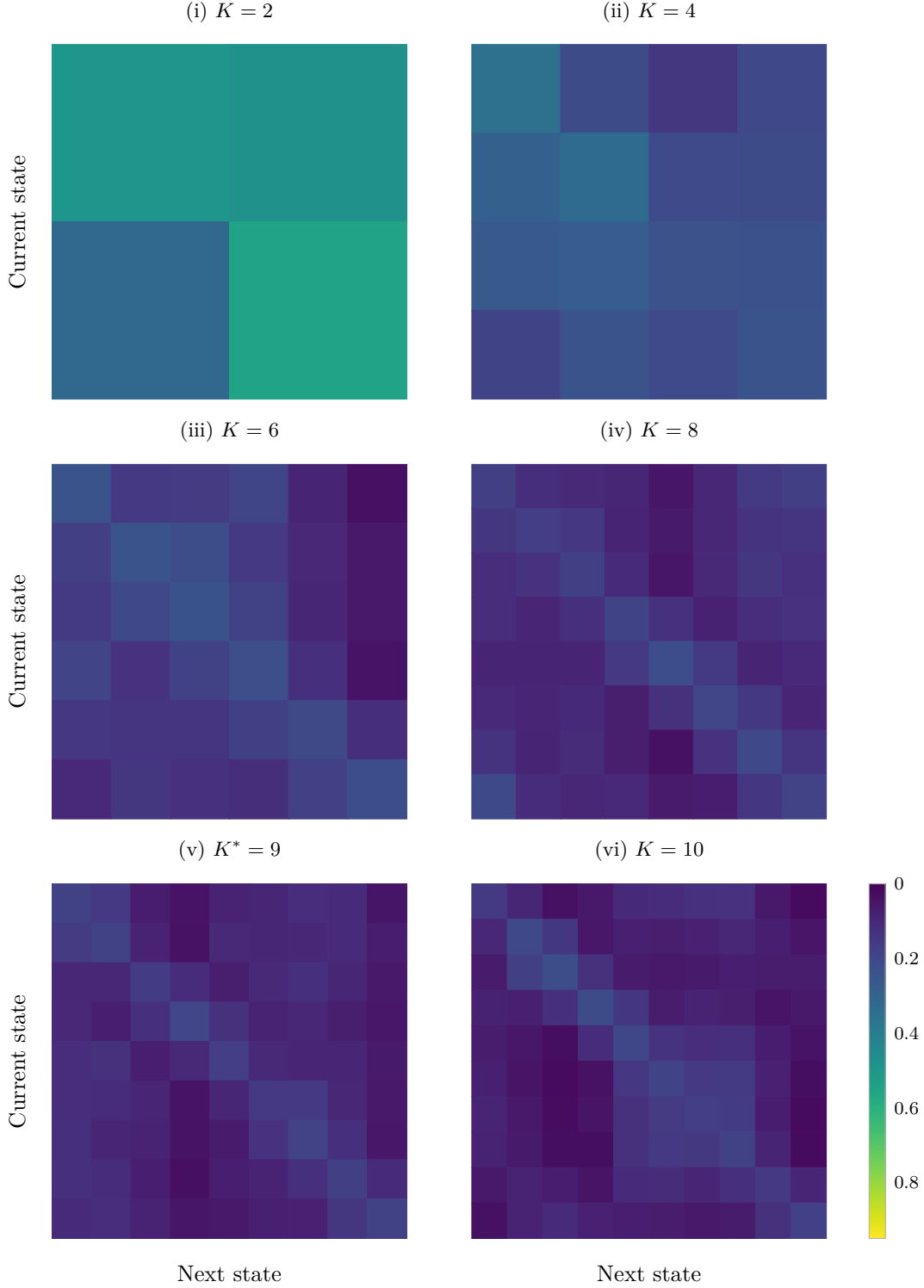
\begin{figure}[H]
    \centering
    \begin{tikzpicture}
        \begin{groupplot}[
                group style={group size=2 by 3, horizontal sep=1cm, vertical sep=1cm,
                        x descriptions at=edge bottom, y descriptions at=edge left},
                width=0.33\textwidth, height=0.33\textwidth, enlargelimits=false,
                axis line style={draw=black!30}, scale only axis,
                xlabel={Next state}, ylabel={Current state},
                xlabel shift=8pt, ylabel shift=8pt,
                xlabel style={font=\small}, ylabel style={font=\small},
                xtick=\empty, ytick=\empty, tick style={draw=none},
                title style={font=\footnotesize}, y dir=reverse,
                colormap/viridis, point meta min=0, point meta max=0.95,
            ]
            \nextgroupplot[title={(i) $K=2$}]
            \addplot[matrix plot*, mesh/cols=2, shader=flat, point meta=explicit] coordinates{
                    (1,1)[0.523](2,1)[0.477](1,2)[0.270](2,2)[0.730]
                };
            \nextgroupplot[title={(ii) $K=4$}]
            \addplot[matrix plot*, mesh/cols=4, shader=flat, point meta=explicit] coordinates{
                    (1,1)[0.548](2,1)[0.197](3,1)[0.051](4,1)[0.204]
                    (1,2)[0.122](2,2)[0.547](3,2)[0.069](4,2)[0.261]
                    (1,3)[0.167](2,3)[0.316](3,3)[0.378](4,3)[0.139]
                    (1,4)[0.192](2,4)[0.370](3,4)[0.041](4,4)[0.398]
                };
            \nextgroupplot[title={(iii) $K=6$}]
            \addplot[matrix plot*, mesh/cols=6, shader=flat, point meta=explicit] coordinates{
                    (1,1)[0.312](2,1)[0.143](3,1)[0.074](4,1)[0.261](5,1)[0.174](6,1)[0.035]
                    (1,2)[0.165](2,2)[0.350](3,2)[0.063](4,2)[0.254](5,2)[0.087](6,2)[0.082]
                    (1,3)[0.110](2,3)[0.081](3,3)[0.467](4,3)[0.096](5,3)[0.193](6,3)[0.055]
                    (1,4)[0.234](2,4)[0.217](3,4)[0.058](4,4)[0.337](5,4)[0.114](6,4)[0.041]
                    (1,5)[0.215](2,5)[0.095](3,5)[0.176](4,5)[0.158](5,5)[0.306](6,5)[0.049]
                    (1,6)[0.105](2,6)[0.187](3,6)[0.107](4,6)[0.130](5,6)[0.121](6,6)[0.351]
                };
            \nextgroupplot[title={(iv) $K=8$}]
            \addplot[matrix plot*, mesh/cols=8, shader=flat, point meta=explicit] coordinates{
                    (1,1)[0.261](2,1)[0.063](3,1)[0.101](4,1)[0.101](5,1)[0.038](6,1)[0.036](7,1)[0.226](8,1)[0.174]
                    (1,2)[0.120](2,2)[0.264](3,2)[0.067](4,2)[0.161](5,2)[0.097](6,2)[0.047](7,2)[0.104](8,2)[0.141]
                    (1,3)[0.149](2,3)[0.053](3,3)[0.303](4,3)[0.071](5,3)[0.048](6,3)[0.069](7,3)[0.195](8,3)[0.111]
                    (1,4)[0.147](2,4)[0.126](3,4)[0.065](4,4)[0.277](5,4)[0.048](6,4)[0.052](7,4)[0.115](8,4)[0.170]
                    (1,5)[0.095](2,5)[0.114](3,5)[0.082](4,5)[0.082](5,5)[0.357](6,5)[0.075](7,5)[0.109](8,5)[0.085]
                    (1,6)[0.091](2,6)[0.070](3,6)[0.109](4,6)[0.102](5,6)[0.084](6,6)[0.358](7,6)[0.103](8,6)[0.083]
                    (1,7)[0.210](2,7)[0.052](3,7)[0.151](4,7)[0.076](5,7)[0.043](6,7)[0.039](7,7)[0.293](8,7)[0.137]
                    (1,8)[0.208](2,8)[0.088](3,8)[0.088](4,8)[0.148](5,8)[0.039](6,8)[0.040](7,8)[0.165](8,8)[0.225]
                };
            \nextgroupplot[title={(v) $K^*=9$}]
            \addplot[matrix plot*, mesh/cols=9, shader=flat, point meta=explicit] coordinates{
                    (1,1)[0.264](2,1)[0.053](3,1)[0.138](4,1)[0.029](5,1)[0.042](6,1)[0.089](7,1)[0.129](8,1)[0.205](9,1)[0.050]
                    (1,2)[0.086](2,2)[0.327](3,2)[0.105](4,2)[0.015](5,2)[0.102](6,2)[0.130](7,2)[0.046](8,2)[0.098](9,2)[0.091]
                    (1,3)[0.157](2,3)[0.076](3,3)[0.234](4,3)[0.014](5,3)[0.037](6,3)[0.159](7,3)[0.059](8,3)[0.207](9,3)[0.056]
                    (1,4)[0.132](2,4)[0.050](3,4)[0.055](4,4)[0.362](5,4)[0.061](6,4)[0.048](7,4)[0.161](8,4)[0.088](9,4)[0.044]
                    (1,5)[0.098](2,5)[0.139](3,5)[0.076](4,5)[0.030](5,5)[0.341](6,5)[0.076](7,5)[0.071](8,5)[0.094](9,5)[0.076]
                    (1,6)[0.123](2,6)[0.119](3,6)[0.195](4,6)[0.013](5,6)[0.046](6,6)[0.228](7,6)[0.054](8,6)[0.162](9,6)[0.059]
                    (1,7)[0.179](2,7)[0.057](3,7)[0.097](4,7)[0.076](5,7)[0.051](6,7)[0.070](7,7)[0.278](8,7)[0.136](9,7)[0.055]
                    (1,8)[0.207](2,8)[0.060](3,8)[0.175](4,8)[0.017](5,8)[0.038](6,8)[0.111](7,8)[0.081](8,8)[0.258](9,8)[0.053]
                    (1,9)[0.113](2,9)[0.130](3,9)[0.106](4,9)[0.020](5,9)[0.070](6,9)[0.094](7,9)[0.067](8,9)[0.116](9,9)[0.286]
                };
            \nextgroupplot[title={(vi) $K=10$}, colorbar,
                colorbar style={at={(1.12,0.5)}, anchor=west, ytick={0,0.2,0.4,0.6,0.8},
                        yticklabel style={font=\scriptsize}, width=0.25cm, height=0.33\textwidth}]
            \addplot[matrix plot*, mesh/cols=10, shader=flat, point meta=explicit] coordinates{
                    (1,1)[0.206](2,1)[0.078](3,1)[0.017](4,1)[0.023](5,1)[0.061](6,1)[0.193](7,1)[0.098](8,1)[0.191](9,1)[0.106](10,1)[0.026]
                    (1,2)[0.078](2,2)[0.274](3,2)[0.033](4,2)[0.088](5,2)[0.061](6,2)[0.121](7,2)[0.061](8,2)[0.181](9,2)[0.067](10,2)[0.036]
                    (1,3)[0.018](2,3)[0.036](3,3)[0.474](4,3)[0.010](5,3)[0.070](6,3)[0.075](7,3)[0.046](8,3)[0.054](9,3)[0.106](10,3)[0.110]
                    (1,4)[0.050](2,4)[0.171](3,4)[0.028](4,4)[0.385](5,4)[0.064](6,4)[0.062](7,4)[0.052](8,4)[0.110](9,4)[0.042](10,4)[0.036]
                    (1,5)[0.059](2,5)[0.067](3,5)[0.063](4,5)[0.030](5,5)[0.386](6,5)[0.081](7,5)[0.096](8,5)[0.097](9,5)[0.054](10,5)[0.066]
                    (1,6)[0.100](2,6)[0.064](3,6)[0.030](4,6)[0.013](5,6)[0.039](6,6)[0.295](7,6)[0.087](8,6)[0.205](9,6)[0.148](10,6)[0.019]
                    (1,7)[0.097](2,7)[0.063](3,7)[0.038](4,7)[0.021](5,7)[0.095](6,7)[0.171](7,7)[0.204](8,7)[0.133](9,7)[0.148](10,7)[0.030]
                    (1,8)[0.102](2,8)[0.117](3,8)[0.024](4,8)[0.026](5,8)[0.051](6,8)[0.207](7,8)[0.071](8,8)[0.279](9,8)[0.099](10,8)[0.023]
                    (1,9)[0.086](2,9)[0.053](3,9)[0.066](4,9)[0.014](5,9)[0.040](6,9)[0.229](7,9)[0.117](8,9)[0.144](9,9)[0.222](10,9)[0.029]
                    (1,10)[0.040](2,10)[0.070](3,10)[0.173](4,10)[0.031](5,10)[0.109](6,10)[0.067](7,10)[0.052](8,10)[0.078](9,10)[0.070](10,10)[0.310]
                };
        \end{groupplot}
    \end{tikzpicture}
    \caption{Row-normalised transition matrices for the Neural VBEM HMM
        ($d_z=64$), engineered features, $K\in\{2,4,6,8,9,10\}$. Panel (v) is
        the selected order $K^*=9$; colour scale identical to
        \cref{fig:appc-bw-transitions,fig:appc-vbem-transitions}.}
    \label{fig:appc-nvbem-transitions}
\end{figure}

The Neural VBEM departs from both raw-feature tiers immediately, at
$K=2$: $A_{1,1}=0.523$, $A_{2,2}=0.730$, forming a far more balanced split than
either Baum-Welch's or VBEM's near-saturating asymmetric pair. At $K^*=9$, the fraud state
(state 4) has $A_{4,4}=0.362$: the highest diagonal entry in this
panel, so still the tier's own most persistent regime, but lower in
absolute terms than either raw-feature tier's fraud-state stickiness
($0.411$, $0.519$). This is not a weaker fraud signal, as the Neural VBEM
fraud state is far more concentrated in fraud content ($50.9\%$ against
$19.4\%$ and $25.3\%$), so a lower self-transition probability still
carries a more discriminative signal per visit. On the joint measure
that matters operationally, stickiness combined with purity,
the Neural VBEM's fraud state is the strongest of the three tiers.

The VBEM and Neural VBEM tiers show the same qualitative pattern as
Baum-Welch at their own selected orders. All three converge on the same
finding by a different route: the fraud-associated regime is persistent
enough to carry real operational signal, but never the single most
persistent regime in the model.

\section{Additional Diagnostics}
\label{app:additional-diagnostics}

State stickiness and transition structure are covered together with the
transition-matrix figures in \cref{app:stickiness}. This section covers the processed-versus-
engineered feature comparison referenced throughout the main text.

\begin{table}[H]
    \centering
    \small
    \setstretch{1.3}
    \setlength{\aboverulesep}{0.7ex}
    \setlength{\belowrulesep}{0.7ex}
    \setlength{\tabcolsep}{6pt}
    \caption{Fraud-state diagnostics under processed features, by tier, at
        the model order given ($K$ is the eligible/selected order where one
        exists).}
    \label{tab:appd-processed}
    \begin{tabular}{lcccc}
        \toprule
        \textbf{Model}          & $K$ & \textbf{Occupancy} & \textbf{Fraud rate} & \textbf{Enrichment} \\
        \midrule
        Baum-Welch HMM          & 8   & $4.3\%$            & $22.3\%$            & $6.31\times$        \\
        VBEM HMM                & -   & -                  & -                   & -                   \\
        Neural VBEM ($d_z=32$)  & 8   & $4.4\%$            & $42.5\%$            & $12.03\times$       \\
        Neural VBEM ($d_z=64$)  & 10  & $3.4\%$            & $48.1\%$            & $13.61\times$       \\
        Neural VBEM ($d_z=128$) & 10  & $4.6\%$            & $30.3\%$            & $8.59\times$        \\
        \bottomrule
    \end{tabular}
\end{table}

\begin{table}[H]
    \centering
    \small
    \setstretch{1.3}
    \setlength{\aboverulesep}{0.7ex}
    \setlength{\belowrulesep}{0.7ex}
    \setlength{\tabcolsep}{6pt}
    \caption{Fraud-state diagnostics under engineered features, by tier, at
        each tier's selected $K^*$ (\cref{tab:model-selection-summary}).}
    \label{tab:appd-engineered}
    \begin{tabular}{lcccc}
        \toprule
        \textbf{Model}          & $K^*$ & \textbf{Occupancy} & \textbf{Fraud rate} & \textbf{Enrichment}    \\
        \midrule
        Baum-Welch HMM          & 8     & $1.8\%$            & $19.4\%$            & $5.48\times$           \\
        VBEM HMM                & 7     & $3.8\%$            & $25.3\%$            & $7.16\times$           \\
        Neural VBEM ($d_z=32$)  & 8     & $3.7\%$            & $46.1\%$            & $13.05\times$          \\
        Neural VBEM ($d_z=64$)  & 9     & $3.8\%$            & $50.9\%$            & $\mathbf{14.43\times}$ \\
        Neural VBEM ($d_z=128$) & 9     & $3.8\%$            & $55.1\%$            & $15.59\times$          \\
        \bottomrule
    \end{tabular}
\end{table}

For the Neural VBEM tier, engineered features win on both fraud rate and
enrichment at every tested $d_z$, by a widening enrichment margin as
$d_z$ grows ($+1.02\times$ at $d_z=32$, $+0.82\times$ at $d_z=64$,
$+6.9\times$ at $d_z=128$), while occupancy stays close between the two
feature sets throughout ($\pm 1$ percentage point). For Baum-Welch, raw peak enrichment across all
$K$ is higher for processed features ($6.31\times$ at $K=8$) than for
engineered's selected value ($5.48\times$, also $K=8$), but
engineered's own peak reaches $7.60\times$ at $K=9$, a $K$ that fails the
emission/posterior agreement criterion and is therefore excluded from
selection under this tier's enforcement of that criterion
(\cref{sec:model-selection}).

For VBEM, processed features fail the agreement criterion at
every candidate $K$ because the emission-based heuristic (highest posterior mean on log
transaction amount) requires the engineered representation's
customer-relative deviation features to be a meaningful fraud proxy at
all, as on raw processed features, a high mean amount is as consistent with
routine high-value legitimate spending as with fraud, so the criterion
is reported but not enforced for that column, and no processed $K^*$ is
selected for VBEM in this study. Engineered features are therefore used
throughout the main because for two of the three tiers the alternative
wins only on a $K$ the selection procedure itself
accepts.

\section{Implementation Details}
\label{app:implementation}

\begin{table}[H]
    \centering
    \small
    \setstretch{1.3}
    \setlength{\aboverulesep}{0.7ex}
    \setlength{\belowrulesep}{0.7ex}
    \setlength{\tabcolsep}{6pt}
    \caption{Encoder architecture and training hyperparameters, Neural VBEM HMM.}
    \label{tab:appf-hyperparameters}
    \begin{tabular}{ll}
        \toprule
        \textbf{Component}                    & \textbf{Value}                                           \\
        \midrule
        \multicolumn{2}{l}{\textbf{Encoder architecture}}                                                \\
        Hidden dimension $d_h$                & 512 (fixed)                                              \\
        Latent dimension $d_z$                & $\{32, 64, 128\}$ (swept)                                \\
        Categorical embedding dim.\ $d_{e_j}$ & $\min(50, \max(2, \lfloor(C_j+1)/2\rfloor))$             \\
        Continuous pathway                    & LayerNorm $\to$ 2-layer MLP, GELU                        \\
        Dropout                               & 0.1                                                      \\
        Fusion / projection                   & LayerNorm $\to$ GELU $\to$ linear, to $\mathbb{R}^{d_z}$ \\
        \midrule
        \multicolumn{2}{l}{\textbf{Phase 1: supervised encoder pretraining}}                             \\
        Objective                             & class-weighted BCE, $\omega = N_-/N_+ \approx 27.6$      \\
        Optimiser                             & Adam, lr $= 10^{-3}$                                     \\
        Epochs (max)                          & 200, early stopping, patience 15                         \\
        Batch size                            & 512                                                      \\
        Gradient clipping                     & $\Vert\nabla\Vert \leq 1.0$                              \\
        \midrule
        \multicolumn{2}{l}{\textbf{Phase 2: VBEM on frozen embeddings}}                                  \\
        NIW posterior init.\                  & $k$-means on a subsample of pretrained embeddings        \\
        Max.\ iterations                      & 500, tolerance $10^{-3}$                                 \\
        Random restarts                       & 3                                                        \\
        Candidate model orders                & $K \in \{2,\ldots,10\}$                                  \\
        \midrule
        \multicolumn{2}{l}{\textbf{Data split}}                                                          \\
        Validation split (held-out log-lik.)  & $15\%$ of training sequences, customer-level             \\
        \midrule
        \multicolumn{2}{l}{\textbf{Default NIW prior} ($d_z$-dependent)}                                 \\
        $\mathbf{m}_0$                        & $\mathbf{0} \in \mathbb{R}^{d_z}$                        \\
        $\kappa_0$                            & 5.0                                                      \\
        $\nu_0$                               & $d_z + 20$                                               \\
        $\mathbf{S}_0$                        & $(d_z+20)\, I_{d_z}$                                     \\
        \bottomrule
    \end{tabular}
\end{table}

The Baum--Welch and raw VBEM scripts use 100 iterations, tolerance
$10^{-3}$, and five restarts across $K\in\{2,\ldots,10\}$. Neural VBEM
uses 500 iterations and three restarts; restart seeds are $137r$ for
$r=0,1,\ldots$. The split and encoder seeds are 42. Both Bayesian HMMs
use initial-state Dirichlet entries 1 and transition-prior entries 1 off
the diagonal and 5 on it; raw categorical-emission concentrations are 1.
The raw Gaussian prior has $m_0=0$, $\kappa_0=1$, $\nu_0=D+1$ and $S_0=I$.
The neural prior is listed above. For the actual diagonal emission
calculations, $a_{0d}=\nu_0/2$ and $b_{0d}=S_{0,dd}/2$.

Encoder pretraining uses Adam and a plateau scheduler that halves the
learning rate after five non-improving epochs, down to $10^{-6}$;
early stopping uses patience 15 and improvement threshold $10^{-5}$.
The source's extra pretraining split is transaction-level, whereas the
outer neural validation split is group-level. Preprocessing and saved
artifact issues are documented in Section~\ref{sec:data-scope}. Some script
filenames differ from the supplied model artifact names, so a versioned
run manifest and regenerated outputs are required for a clean rerun.

\end{document}